\documentclass{article}

\usepackage{times}
\usepackage{graphicx} % more modern
\usepackage{subfigure}

\usepackage{natbib}

\usepackage[algo2e,ruled,vlined]{algorithm2e}

\usepackage{hyperref}

\usepackage[accepted]{icml2016}

\icmltitlerunning{Towards Practical Bayesian Parameter and State Estimation}

\usepackage{amsthm}
\usepackage{amsfonts}
\usepackage{amsmath}
\usepackage{amssymb}
\usepackage{xspace}

\usepackage{listings}
\usepackage{paralist}

\newcommand{\ben}{\begin{enumerate*}}
\newcommand{\een}{\end{enumerate*}}
\newcommand{\beq}{\begin{eqnarray}}
\newcommand{\eeq}{\end{eqnarray}}
\newcommand{\bit}{\begin{itemize*}}
\newcommand{\eit}{\end{itemize*}}
\newcommand{\nn}{\nonumber}
\newcommand{\hide}[1]{}
\newcommand{\indep}[1]{\perp}

\newcommand{\gauss}{\ensuremath{\mathcal{N}}}
\newcommand{\filter}{API\xspace}
\newcommand{\name}{SPEC\xspace}
\renewcommand{\cite}{\citep}
\newcommand{\sinmodel}{\texttt{SIN}\xspace}
\newcommand{\birdmodel}{\texttt{BIRD}\xspace}
\newcommand{\slammodel}{\texttt{SLAM}\xspace}
\begin{document}

\twocolumn[
\icmltitle{Towards Practical Bayesian Parameter and State Estimation}

% It is OKAY to include author information, even for blind
% submissions: the style file will automatically remove it for you
% unless you've provided the [accepted] option to the icml2016
% package.
\icmlauthor{Yusuf B. Erol}{yusufbugraerol@berkeley.edu}
\icmlauthor{Yi Wu}{jxwuyi@cs.berkeley.edu}
\icmladdress{UC Berkeley}
\icmlauthor{Lei Li}{lileicc@gmail.com}
\icmladdress{Baidu Research}
\icmlauthor{Stuart Russell}{russell@cs.berkeley.edu}
\icmladdress{UC Berkeley}

% You may provide any keywords that you
% find helpful for describing your paper; these are used to populate
% the "keywords" metadata in the PDF but will not be shown in the document
\icmlkeywords{boring formatting information, machine learning, ICML}

\vskip 0.3in
]

\begin{abstract}
%\reminder{to rewrite the abstract}
Joint state and parameter estimation is a core problem for dynamic Bayesian networks. 
Although modern  probabilistic inference toolkits make it relatively easy to specify large and practically relevant probabilistic models, the silver bullet---an efficient and general online inference algorithm for such problems---remains elusive, forcing users to write special-purpose code for each application.
We propose a novel blackbox algorithm -- a hybrid of particle filtering for state variables and assumed density filtering for parameter variables.  
It has following advantages: 
(a) it is efficient due to its online nature, and 
(b) it is applicable to both discrete and continuous parameter spaces .
On a variety of toy and real models, our system is able to generate more accurate results within a fixed computation budget. 
This preliminary evidence indicates that the proposed approach is likely to be of practical use.
%Rather than wait for the silver bullet, we propose instead a practical engineering solution that combines a reasonably general and effective algorithm (a hybrid of particle filtering for state variables with assumed density filtering for parameter variables) with modern compiler technology that generates extremely efficient, model-specific inference code automatically. We demonstrate this approach in the context of the BLOG PPL, which provides an elegant modelling language for complex probability models. We encode three very different temporal models with unknown parameters and show that the generated code is 2--3 orders of magnitude more efficient than other state-of-the-art tools as well as providing far more accurate results for a fixed computational budget. This preliminary evidence indicates that the proposed approach is likely to be of practical use.
\end{abstract}

% Probabilistic programming languages (PPL) aim to provide 
% powerful representations for probabilistic models. 
% Ideally, PPL models are solved by generic inference engines such that 
% domain experts could focus on model building. 
% While such PPL systems exist for Bayesian networks, it has been an open challenge to create an automatic and effective inference engine for dynamical Bayesian networks (DBNs). 
% In this paper, we present a compiled inference system, \name, for inference on practical DBNs. \name utilizes multiple novel compiler level optimizations as well as a new black box algorithm, \filter, for the core inference problem: jointly estimate state (dynamic variables) and parameters (static variables), no matter whether their values are discrete or continuous. 
% We evaluated both the statistical and algorithmic performance on several examples, including a practical application with real world data. 
% Our experiments show that our system (a) is able to handle a general class of models and improve the efficiency by orders of magnitudes; (b) even using much less computational resources, converges to accurate approximations much more quickly compared to existing ones.
% \reminder{The selling point is that we have a practical system. The system has two properties: (a) ... (b) ...}
% \end{abstract}

\section{Introduction}\label{sec:intro}
% \reminder{
% 1. PPL is useful, One real need is to handle dynamic models. We need generic inference engine. 
% 2. For dynamic models in PPL, it is a challenge to joint estimate state and parameter. Existing generic solution particle filters do not work, because ..
% 3. goal of this paper. contribution: fast practical system. The algorithmic problem of the core system is 
% ... , brief summary of experiment findings.

% 1.1 related work
% existing PPLs, and PP for dyanamic model, current limitation. 
% inference algorithms for join ...: particle ...
% Why fail.

% from stuart:
% 1. state parameter problem importatnt
% 2. in need of generic  inference alg. 
% 3. PPL provides an expressive way ...
% 4. in order to solve practical problem, often long duration and large number of particles, need efficient inference engine.

% }

% Probabilistic programming languages (PPLs) are a very promising approach
% for solving this long-standing problem in AI, machine learning, and statistics: the provision of an expressive, general-purpose modeling language capable of
% handling uncertainty, combined with a general-purpose inference engine
% able to handle any model the user might construct. The user (and, a fortiori,
% the brain) should not be required to carry out machine learning research and implement new algorithms for each problem that comes along.

Many problems in scientific studies and natural real-world applications involve modelling of 
dynamic processes, which are often modeled by dynamic probabilistic models (DPM)~\cite{elmohamed2007collective,arora2010global}. Online state and parameter estimation --computing the posterior probability for both (dynamic) states and (static) parameters, incrementally over time-- is crucial for many applications such as simultaneous localization and mapping, real-time clinical decision support, and process control.

Various sequential Monte-Carlo state estimation methods have been introduced for real-world applications  \cite{gordon1993novel,arulampalam2002tutorial,cappe2007overview,lopes2011particle}. It is yet a challenge to estimate parameters and state jointly for DPMs with complex dependencies and nonlinear dynamics. 
Real-world models can involve both discrete and continuous variables, arbitrary dependencies and a rich collection of nonlinearities and distributions. 
Existing algorithms either apply to a restricted class of models~\cite{storvik2002particle}, 
or are very expensive in time complexity~\cite{andrieu2010particle}. 
%\reminder{do we need this last sentence? also grammar is off}
%Previously to turn those algorithms into a practical system one needs create a special purpose solution for every new application, which requires huge engineering efforts. 

DPMs in real-world applications often have a large number of observations and one will need a large number of particles to perform the inferential task accurately, which requires the inference engine to be efficient and scalable. 
%Recently emerging probabilistic programming languages (PPLs) enable convenient description of such probabilistic models~\cite{milch2005blog,pfeffer2009figaro}. 
%In addition, PPLs aim to provide automatic inference tools for any models that developers might construct in the languages. 
While there is much success in creating generic inference algorithms and learning systems for
non-dynamical models, it remains open for DPMs. 
A black-box inference system for DPMs needs two elements to be practically useful: a general and effective joint state/parameter estimation algorithm and an efficient implementation of inference engine. 
In this paper, we propose a practical online solution for the general combined state and parameter estimation problem in DPMs.
We developed State and Parameter Estimation Compiler (\name) for arbitrary DPMs described in a declarative modelling language BLOG~\cite{milch2005blog}. 
\name is equipped with a new black box inference algorithm named Assumed Parameter Inference (\filter)
which is an hybrid of particle filtering for state variables with assumed density filtering for parameter variables. \name is also geared with an optimizing compiler to generate application specific inference code for efficient inference on real-world applications. 
%The system utilizes multiple modern compilation techniques for efficient inference on real-world applications. 

Our contribution of the paper is as follows: 
\begin{inparaenum}[(1)]
\item We proposed a new general online algorithm, \filter, for joint state and parameter estimation, regardless whether the parameters are discrete or continuous, and whether the dynamics is linear or nonlinear;
%\item We proposed compilation techniques to speedup the inference in practical scale applications;
\item We developed a black box inference engine, \name, equipped with \filter for arbitrary DPMs.  \name also utilizes modern compilation techniques to speedup the inference for practical applications;
\item We conducted experiments and demonstrated \name's superior performance in terms of both accuracy and efficiency on several real-world problems. 
%\name achieves over two orders of magnitudes speedup in running time and produces significantly more accurate estimation results within a fixed amount of time limit.
\end{inparaenum}

We organize the paper as follows: Section \ref{sec:background} reviews the state and parameter estimation literature, section \ref{sec:algo} describes our new algorithm \filter, and introduces our black box inference engine \name, and section \ref{sec:experiment} provides experiment results. Further discussion is given in section \ref{sec:related}. Finally we conclude our paper in section \ref{sec:conclude}.

\section{Background}\label{sec:background}

%\reminder{This whole section needs revision or re-organized}

%\textbf{Existing algorithms for state and parameter estimation: }
%\reminder{think about moving some of the following content to related work section}
% In the context of combined state and parameter estimation, unknown model parameters may be represented by static parameters that define the transition and observation probabilities for the Markov process. Particle filters fail for such models as the particles are incapable of exploring the parameter space: at initialization the particle filter samples parameter values from the prior, but these values remain fixed for the static parameter; and over time, the resampling process removes all but one particle which is known as \emph{sample impoverishment}. 
An overview of SMC methods for parameter estimation is provided in \cite{kantas2015particle}. 
Plain particle filter fails to estimate parameters due to the inability to explore the parameter space. This problem is particularly severe in high-dimensional parameter spaces, as a particle filter would require exponentially many (in the dimensionally of the parameter space) particles to sufficiently explore the parameter space. 
Various algorithms have been proposed to achieve the combined state and parameter estimation task, however, the issue still remains open as existing algorithms either suffer from bias or computational efficiency. The \emph{artificial dynamics} approach \cite{liu2001combined}, although computationally efficient and applicable to arbitrary continuous parameter models, results in biased estimates and fails for intricate models considered in this paper.  The resample-move algorithm \cite{gilks2001following} utilizes kernel moves that target $p(\theta,x_{0:t} \mid y_{0:t})$ as its invariant. However, this method requires $O(t)$ computation per time step, leading  \citeauthor{gilks2001following} to propose a move at a rate proportional to $1/t$ so as to have asymptotically constant-time updates. \citet{fearnhead2002markov}, \citet{storvik2002particle} and \citet{lopes2010particle} proposed sampling from $p(\theta \mid x_{0:t}, y_{0:t})$ at each time step for models with fixed dimensional sufficient statistics. However, arbitrary models generally do not accept sufficient statistics for $\theta$ and for models with sufficient statistics, these algorithms require the sufficient statistics to be explicitly defined by the user. The extended parameter filter \cite{erol2013extended} generates approximate sufficient statistics via polynomial approximation, however, it requires hand-crafted manual approximations. The gold-standard particle Markov Chain Monte Carlo (PMCMC) sampler introduced by \citet{andrieu2010particle} converges to the true posterior and is suitable for building a black box inference engine (i.e., LibBi~\cite{murray2013bayesian} and Biips~\cite{todeschini2014biips}). With the black-box engine, a user can purely work on the machine learning research without worrying about the implementation of algorithms for each problem that comes along.
However, note that PMCMC is an offline algorithm, which is unsuitable for real-time applications. Moreover, it may have poor mixing properties which in turn necessitates launching the filtering processes substantially many times, which can be extremely expensive for real-world applications with large amount of data.

% EM approaches \cite{schon2011system,andrieu2003online} provide only point estimates and tend to get stuck at local optima. \citet{poyiadjis2011particle} approximated the score and the information matrix in order to optimize the log-likelihood of the observed data. However, this method provides point estimates and requires computation of gradients and Hessians which is not available for general state-space models.
% We are striving for an algorithm that bridges the gap between general algorithms like the Liu-West filter and \mbox{PMCMC} which do not use any information specific to the model and algorithms that relies on model-specific information like sufficient statistics or gradients to fit into the \emph{probabilistic programming} framework \cite{gordon2014probabilistic}.
\hide{
\textbf{Existing PPLs for state and parameter estimation: }
% There are many well-known PPLs and have been a number
% of successes using PPLs for real applications such as entity
% resolution~\cite{singla2006entity}, citation
% matching~\cite{pasula2002identity}, relation
% extraction~\cite{yoshikawa2009jointly} and seismic
% monitoring~\cite{arora2010global}.

Several PPLs and PP systems provide particle filtering engines for dynamical probabilistic models (DPMs), including universal PPLs systems such as BLOG~\cite{milch2005blog} and Figaro~\cite{pfeffer2009figaro} as well as the specialized system Biips~\cite{todeschini2014biips}. These systems provides flexibility for user to express the DPMs in a high level modelling language without interacting with the backend inference system.
Other PPLs such as Stan~\cite{sdt2014stan} and Church~\cite{goodman2008church} have support for limited model class.

However, towards practical situations, existing systems either produce inaccurate estimations or need too much time before the algorithm converges. For the latter case, it might be simply because the system executes the algorithm too inefficiently.
In the system perspective, a practical inference engine in a PPL for DPMs should achieve the following properties: (1) great expressiveness power for modelling real-world applications; (2) a compiled back-end to get rid of interpretation overhead and to leverage the power of compiler level optimizations; (3) both memory and computational efficiency in capable of handling a large number of particles.

Among our benchmark systems, BLOG, Figaro and Biips, BLOG and Figaro are Turing-complete for their expressiveness power but they suffer from tremendous interpretation overhead by interpreting the inference algorithm over a complicated internal data structure. Biips is another embedded system specialized for DPMs with very restricted syntax (i.e., Biips cannot express any mixtures of distributions). It is integrated in R and Matlab and also suffers from significant memory management overhead. 
}

\section{Assumed Parameter Inference (API)}\label{sec:algo}
Let $\Theta$ be a parameter space for a partially observable
Markov process $\left\{X_{t} \right\}_{t \geq 0}, \left \{Y_t \right \}_{t \geq
0}$  \mbox{which is defined as follows:}
{
\small
\begin{align}
	X_0 & \sim p(x_0) \\
    X_t \mid x_{t-1} &\sim p(x_t \mid x_{t-1}, \theta) \\
	Y_t \mid x_t &\sim p(y_t \mid x_t, \theta)
\label{eq:state-space-model}
\end{align} 
} 
Here the state variables $X_t$ are unobserved and the observations $Y_{t}$ are assumed to be conditionally independent of
other observations given $X_{t}$.  We assume in this section that the states $X_t$ and observations $Y_t$ are
 vectors in $d$ and $m$ dimensions respectively. 
The model parameter $\theta$ can be both continuous and discrete. 

% The filtering density $p(x_t \mid y_{0:t}, \theta)$ obeys the following
% recursion:
% \begin{align}
% \label{eq:bayes1}
% &	p(x_t  \mid y_{0:t}, \theta) =\frac{p(y_t \mid x_t, \theta) p(x_t \mid
%     y_{0:t-1}, \theta) }{ p(y_t  \mid y_{0:t-1}, \theta) } \notag \\
%   &=  \frac{p(y_t \mid x_t, \theta)  }{ p(y_t  \mid y_{0:t-1}, \theta) } \int p(x_{t-1} \mid
%         y_{0:t-1}, \theta) p(x_{t}\mid x_{t-1}, \theta) d x_{t-1}
% \end{align}
% where the update steps for $p(x_t  \mid y_{0:t-1}, \theta)$ and $p(y_t \mid
% y_{0:t-1}, \theta)$ involve the evaluation of integrals that are in general not tractable.

Our algorithm, Assumed Parameter Inference (API) approximates the posterior density $p(x_t, \theta \mid y_{0:t})$ via particles following the framework of sequential Monte-Carlo methods. 
At time step $t$, for each particle path, we sample from $q_t^i(\theta)$ which is the approximate representation of $p(\theta \mid x_{0:t}^i,y_{0:t})$ in some parametric family $\mathcal{Q}$. $N$ particles are used to approximately represent the state and parameters and additional $M$ samples for each particle path are utilized to perform the moment-matching operations required for assumed density approximation as explained in section \ref{sec:adf}.
The proposed method is illustrated in Algorithm \ref{alg:proposed}. Notice that at the propagation step, we are using the bootstrap proposal density, i.e. the transition probability. As in other particle filtering methods, better proposal distributions will improve the performance of the algorithm.

\begin{algorithm2e}[htb]
\label{alg:proposed}
\caption{Assumed Parameter Inference}
\KwIn{$y_{0:T}$, $\mathcal{Q}$, $N$, and $M$,prior}
\KwOut{Samples $\left\{x_{0:T}^i, \theta^i \right\}_{i=1}^N$}
Initialize $\left\{x_0^i,q_0^i(\theta)\right\}_{i=1}^N$ according to the prior\; 
\For{$t=1,\ldots,T$} {
\For{$i=1,\dots,N$}{
	sample $\theta^i \sim q_{t-1}^{i}(\theta) \approx p(\theta \mid x_{0:t-1}^i,y_{0:t-1})$\;
	sample $x_{t}^i \sim p(x_t \mid x_{t-1}^i,\theta^i)$ \;
	$w_t^i \leftarrow p(y_t \mid x_t^i,\theta^i)$\;
	$q_t^i(\theta) \leftarrow \mathrm{Update}(M; q_{t-1}^i(\theta),x_t^i,x_{t-1}^i,y_t)$\;
}
sample $\left\{\frac{1}{N},\bar{{x}}_t^i,\bar{{q}}_t^i \right\}\leftarrow  $Multinomial$\left\{w_t^i,{x}_t^i, q_t^i \right\}$\;
$\left\{{x}_t^i, q_t^i \right\}\leftarrow \left\{\bar{{x}}_t^i, \bar{{q}}_t^i\right\}$\;
}
\end{algorithm2e}

 We are approximating $p(\theta \mid x_{0:t},y_{0:t})$ by exploiting a family of basis distributions. In our algorithm this is expressed through the $\mathrm{Update}$ function. The $\mathrm{Update}$ function generates the approximating density $q$ via minimizing the KL-divergence between the target $\hat{p}$ and the basis $q$.

\subsection{Approximating $p(\theta \mid x_{0:t},y_{0:t})$}
\label{sec:adf}
At each time step with each new incoming data point we approximate the posterior distribution by a tractable and compact distribution from $\mathcal{Q}$. 
Our approach is inspired by assumed density filtering (ADF) for state estimation~\cite{lauritzen1992propagation,boyen1998tractable}. 
%Assumed density filtering (ADF) is a general approximate Bayesian inference technique that sequentially approximates a target density via projection onto the space of desirable distributions. ADF has been reinvented in different communities under different names \cite{lauritzen1992propagation,boyen1998tractable,opper1998bayesian}. ADF also forms the backbone of the expectation-propagation (EP) algorithm. EP has been successfully applied to Bayesian networks \cite{minka2001expectation} and dynamic Bayesian networks \cite{heskes2002expectation}. 

For our application, we are interested in approximately representing $p(\theta \mid x_{0:t},y_{0:t})$ in a compact form that belongs to a family of distributions.
\beq
	p(\theta \mid x_{0:t},y_{0:t}) &\propto&  \prod_{k=0}^t t_k(\theta) \nn \\
	t_k(\theta) &=&  \left\{
	\begin{matrix}
	p(\theta)p(y_0 \mid x_0,\theta), k=0 \\
	p(y_k \mid x_k,\theta) p(x_k \mid x_{k-1},\theta), k \geq 1
	\end{matrix}
	\right. \nn
\eeq
Let us assume that at time step $k-1$ the posterior was approximated by $q_{k-1} \in \mathcal{Q}$. Then,
% where $\mathcal{Q}$ is the family of probability distributions we would like to work with. Then 
\beq
	\hat{p}(\theta \mid x_{0:k},y_{0:k}) = \frac{t_k(\theta)q_{k-1}(\theta)}{\int_{\theta} t_k(\theta)q_{k-1}(\theta) }
\eeq
For most models, $\hat{p}$ will not belong to $\mathcal{Q}$. ADF projects $\hat{p}$ into $\mathcal{Q}$ via minimizing KL-divergence:
\beq
	q_k(\theta) =\arg\min_{q \in \mathcal{Q}} D \left(  \hat{p}(\theta \mid x_{0:k},y_{0:k}) \mid \mid q(\theta) \right)
\eeq
For $\mathcal{Q}$ in the exponential family, minimizing the KL-divergence reduces to moment matching \cite{expectation05seeger}. For $q(\theta) \propto \exp \left\{ < \gamma, f(\theta) > \right\}$, where $f(\theta)$ is the sufficient statistic, the minimizer of the KL-divergence satisfies the following:
\beq
	g(\cdot) = \int f(\theta)q_k(\theta)d\theta &=& \int f(\theta)\hat{p}(\theta) d\theta \nn \\
	&\propto& \int f(\theta)t_k(\theta)q_{k-1}(\theta)d\theta \nn
\eeq		
where $g(\cdot)$ is the link function. Thus, for the exponential family, the $\mathrm{Update}$ function computes the moment matching integrals to update the canonical parameters of $q_k(\theta)$. These integrals are in general intractable. We propose approximating the above integral by a Monte Carlo sum with $M$ samples, sampled from $q_{k-1}(\theta)$ as follows:
% In the following subsection, we will describe a general approximation scheme for the aforementioned integrals.
\label{sec:exp_fam}
% For $q(\theta) \propto \exp \left\{ < \gamma, f(\theta) > \right\}$, where $f(\theta)$ is the sufficient statistic, the moment matching integral to be evaluated is:
% \beq
% 	g(\gamma_k) = \int f(\theta)t_k(\theta)q_{k-1}(\theta)d\theta
% \eeq	
% where $g(\cdot)$ is the link function. We will approximate above integral by a Monte Carlo sum with $M$ samples, sampled from $q_{t-1}(\theta)$.
{
\small
\beq
    Z &\approx& \frac{1}{M}\sum_{j=1}^M t_k(\theta^j) \nn \\
	g(\cdot) &\approx& \frac{1}{MZ} \sum_{j=1}^M f(\theta^j)t_k(\theta^j), \> \mathrm{where} \> \theta^j \sim q_{k-1}(\theta) \nn
\eeq
}
In our framework, this approximation is done for all particle paths $x_{0:k}^i$ and the corresponding $q_{k-1}^i$, hence leading to a total of $O(NM)$ samples.

\subsubsection{Gaussian}
It is worthwhile investigating the Gaussian case. For a multivariate Gaussian $\mathcal{Q}$, explicit recursions can be derived.
\beq 
	\hat{p}(\theta) \propto t_k(\theta) q_{k-1}(\theta)
\eeq
where $q_{k-1}(\theta) = \gauss(\theta; \mu_{k-1},\Sigma_{k-1})$. Then;
{
\small
\beq
\label{eq:moment_match}
	Z &=& \int \hat{p}(\theta)d\theta = \int t_k(\theta)q_{k-1}(\theta) d\theta  \nn \\
	\mu_k &=& \frac{1}{Z}\int \theta t_k(\theta) q_{k-1}(\theta) d\theta  \\
	\Sigma_k &=&  \frac{1}{Z}\int \theta \theta^T t_k(\theta) q_{k-1}(\theta) d\theta -\mu_k \mu_k^T \nn
\eeq
}
These integrals can be approximated via Monte Carlo summation as described in section \ref{sec:exp_fam}. Another alternative is \emph{deterministic sampling}. Since $q$ is multivariate Gaussian, Gaussian quadrature rules can be utilized. In the context of expectation-propagation this has been proposed by \citet{zoeter04a}. In the context of Gaussian filtering, similar quadrature ideas have been applied as well \cite{huber2008gaussian}. 

% Furthermore, the quadrature points need to be sampled once at the initialization of the algorithm. The quadrature points will then be transformed with respect to the mean and the covariance of $q_{k-1}(\theta)$ at each time step (for the one dimensional case this transformation would be $\sqrt{2}\sigma_k r+\mu_k$, where $r$ is the quadrature points sampled at initialization).
For a polynomial $f(x)$ of order up to $2M-1$, $\int f(x)e^{-x^2}dx$ can be calculated exactly via Gauss-Hermite quadrature with $M$ quadrature points. Hence, the required moment matching integrals in Equation \ref{eq:moment_match} can be approximated arbitrarily well by using more quadrature points. The Unscented transform \cite{julier2004unscented} is one specific Gaussian quadrature rule where one would only use $M=2p$ \emph{deterministic} samples for approximating an integral involving a $p$-dimensional multivariate Gaussian. In our case these samples are:
\beq
	\theta_j &=& \mu_{k-1} + \left( \sqrt{p\Sigma_{k-1}} \right)_j, j = 1,\dots,p  \nn \\
	\theta_{p+j} &=&  \mu_{k-1} - \left( \sqrt{p\Sigma_{k-1}} \right)_j, j = 1,\dots,p
\eeq
where $\left( \cdot \right)_j$ means the $j$th column of the corresponding matrix. Then, one can approximately evaluate the moment matching integrals as follows:
{
\small
\beq
	Z &\approx& \frac{1}{2p} \sum_{j=1}^{2p} t_k(\theta^j) \nn \\
	\mu_k &\approx& \frac{1}{2pZ} \sum_{j=1}^{2p} \theta^j t_k(\theta^j) \nn \\
	\Sigma_k &\approx& \frac{1}{2pZ} \sum_{j=1}^{2p} \theta^j (\theta^j)^T t_k(\theta^j) - \mu_k \mu_k^T \nn
\eeq
}
\vspace{-1em}
% The errors induced by these approximations are investigated in section \ref{sec:exp}.
\subsubsection{Mixture of Gaussians}
Weighted sum of Gaussian probability density functions can be used to approximate another density function arbitrarily closely. Mixture of Gaussians has been used in the context of state estimation as early as 1970s \cite{alspach1972nonlinear}. 
%\citet{alspach1972nonlinear} utilized linearization and extended Kalman filter updates to approximate the filtering density via mixtures. \citet{huber2011mixture} used unscented Kalman filter--like updates in order to approximately track the filtering density with mixtures. 

%Let us assume that at time step $k-1$ it was possible to represent $p(\theta \mid x_{0:k-1},y_{0:k-1})$ as a mixture of Gaussians with $L$ components. 
%\beq
%	p(\theta \mid x_{0:k-1},y_{0:k-1}) &=& \sum_{m=1}^{L} \alpha_{k-1}^m \gauss(\theta;\mu_{k-1}^m,\Sigma_{k-1}^m) \nn \\
%				&=& q_{k-1}(\theta) \nn
%\eeq
Let us assume that at time step $k-1$ it was possible to represent $p(\theta \mid x_{0:k-1},y_{0:k-1})$ as a mixture of Gaussians with $L$ components. 
\beq
	p(\theta \mid x_{0:k-1},y_{0:k-1}) &=& \sum_{m=1}^{L} \alpha_{k-1}^m \gauss(\theta;\mu_{k-1}^m,\Sigma_{k-1}^m) \nn \\
				&=& q_{k-1}(\theta) \nn
\eeq
Given new data $x_k$ and $y_k$;
\beq
	\hat{p}(\theta \mid x_{0:k},y_{0:k}) \propto \sum_{m=1}^L \alpha_{k-1}^m t_k(\theta)  \gauss(\theta;\mu_{k-1}^m,\Sigma_{k-1}^m) \nn
\eeq
The above form will not be a Gaussian mixture for arbitrary $t_k$. We can rewrite it as:
\beq
	\hat{p} \propto \sum_{m=1}^L \alpha_{k-1}^m \beta^m \frac{t_k(\theta) \gauss(\theta;\mu_{k-1}^m,\Sigma_{k-1}^m)}{\beta^m}
\eeq	
where the fraction is to be approximated by a Gaussian via moment matching and the weights are to be normalized. Here, each $\beta^m = \int t_k(\theta)  \gauss(\theta;\mu_{k-1}^m,\Sigma_{k-1}^m)d\theta$ describes how well the $m$-th mixture component explains the new data. That is, a mixture component that explains the new data well will get up-weighted and vice versa. The resulting approximated density would be $q_k(\theta) = \sum_{m=1}^K  \alpha_{k}^m \gauss(\theta;\mu_{k}^m,\Sigma_{k}^m)$ where the recursions for updating each term is as follows:
%Given new data $x_k$ and $y_k$; we approximate $p(\theta \mid x_{0:k},y_{0:k})$ with $q_k(\theta)$ according to the following recursions.
%\beq
% 	\hat{p}(\theta \mid x_{0:k},y_{0:k}) \propto \sum_{m=1}^K \alpha_{k-1}^m t_k(\theta)  \gauss(\theta;\mu_{k-1}^m,\Sigma_{k-1}^m) \nn
%\eeq
% The above form will not be a Gaussian mixture for arbitrary $t_k$. We can rewrite it as:
% \beq
% 	\hat{p} \propto \sum_{m=1}^K \alpha_{k-1}^m \beta^m \frac{t_k(\theta) \gauss(\theta;\mu_{k-1}^m,\Sigma_{k-1}^m)}{\beta^m}
% \eeq	
% where the fraction is to be approximated by a Gaussian via moment matching and the weights are to be normalized. Here, each $\beta^m = \int t_k(\theta)  \gauss(\theta;\mu_{k-1}^m,\Sigma_{k-1}^m)d\theta$ describes how well the $m$-th mixture component explains the new data. That is, a mixture component that explains the new data well will get up-weighted and vice versa. The resulting approximated density would be $q_k(\theta) = \sum_{m=1}^K  \alpha_{k}^m \gauss(\theta;\mu_{k}^m,\Sigma_{k}^m)$ where the recursions for updating each term is as follows:
{
\small
\beq
	\beta^m  &=&  \int t_k(\theta) \gauss(\theta; \mu_{k-1}^m, \Sigma_{k-1}^m) d\theta \nn \\
	\alpha_k^m &=& \frac{\alpha_{k-1}^m \beta^m}{\sum_{\ell} \alpha_{k-1}^{\ell} \beta^{\ell} } \nn \\
	\mu_k^m &=& \frac{1}{\beta^m}\int \theta t_k(\theta) \gauss(\theta; \mu_{k-1}^m, \Sigma_{k-1}^m) d\theta \nn \\
	\Sigma_k^m &=& \frac{1}{\beta^m}\int \theta \theta^T t_k(\theta) \gauss(\theta; \mu_{k-1}^m, \Sigma_{k-1}^m) d\theta -\mu_k^m(\mu_k^m)^T \nn
\eeq
} 
%The details of the derivation is presented in the appendix due to the space limit. 
Similar to the Gaussian case, the above integrals are generally intractable. Either a Monte Carlo sum or a Gaussian quadrature rule can be utilized to approximately update the means and covariances. 

\subsubsection{Discrete Parameter Spaces}
Let us consider a $p$-dimensional parameter space where each parameter can take at most $\left| \Theta \right|$ values. For discrete parameter spaces, one can always track \mbox{$p(\theta \mid x_{0:t},y_{0:t})$} in a constant time per update fashion since the posterior will be evaluated at finitely many points. The number of points, however, grows exponentially as $\left| \Theta \right|^p$. Hence, tracking the sufficient statistics becomes computationally intractable with increasing dimensionality. For discrete parameter spaces we propose projection onto a fully factorized distribution, i.e. $q_t(\theta) = \prod q_{i,t}(\theta_i)$.
For this choice, minimizing the KL-divergence reduces to matching marginals.
\beq
	Z &=& \sum_{\theta} t_k(\theta) q_{k-1}(\theta) \nn \\
	q_{i,k}(\theta_i) &=& \frac{1}{Z}  \sum_{\theta \setminus \theta_i} t_k(\theta) q_{k-1}(\theta)
\eeq
Computing these summations is intractable for high-dimensional models, hence we propose using Monte Carlo summation as described in subsection \ref{sec:exp_fam}. In the experiments section, we consider a simultaneous localization and mapping problem where the map is discrete. 

\subsection{Black-Box Inference}
%\reminder{Yi will add a short discussion of how it is possible for us to make black-box inference with API, then shortly introduce the language/SIN model. The memory efficieny, memoization, optimizations etc. will be also discussed. Whenever a property is mentioned, we will also refer to the experiments to validate via empirical results. In section 3.3. we will introduce the complexity of the algorithm and compare against PMCMC.}
%This section contains the discussion, design and implementation of the inference engine.

%We firstly introduce the language interface of Swift, and then introduce the compiler level optimizations for both memory and computational efficiency. We end this section by illustrating the integrated algorithms.

As discussed above, when the form of $q_k$ is fixed, \filter can be performed over any dynamic probabilistic model for which computing and sampling from the $p(y_k|x_k,\theta)$ and $p(x_k|x_{k-1},\theta)$ is viable.

\begin{figure}[htb]
%\center
%{frame=lines,baselinestretch=1,fontsize=\scriptsize,linenos,numbersep=4pt}
%\begin{scriptsize}
%\begin{minted}{scala}
\includegraphics[width=0.45\textwidth]{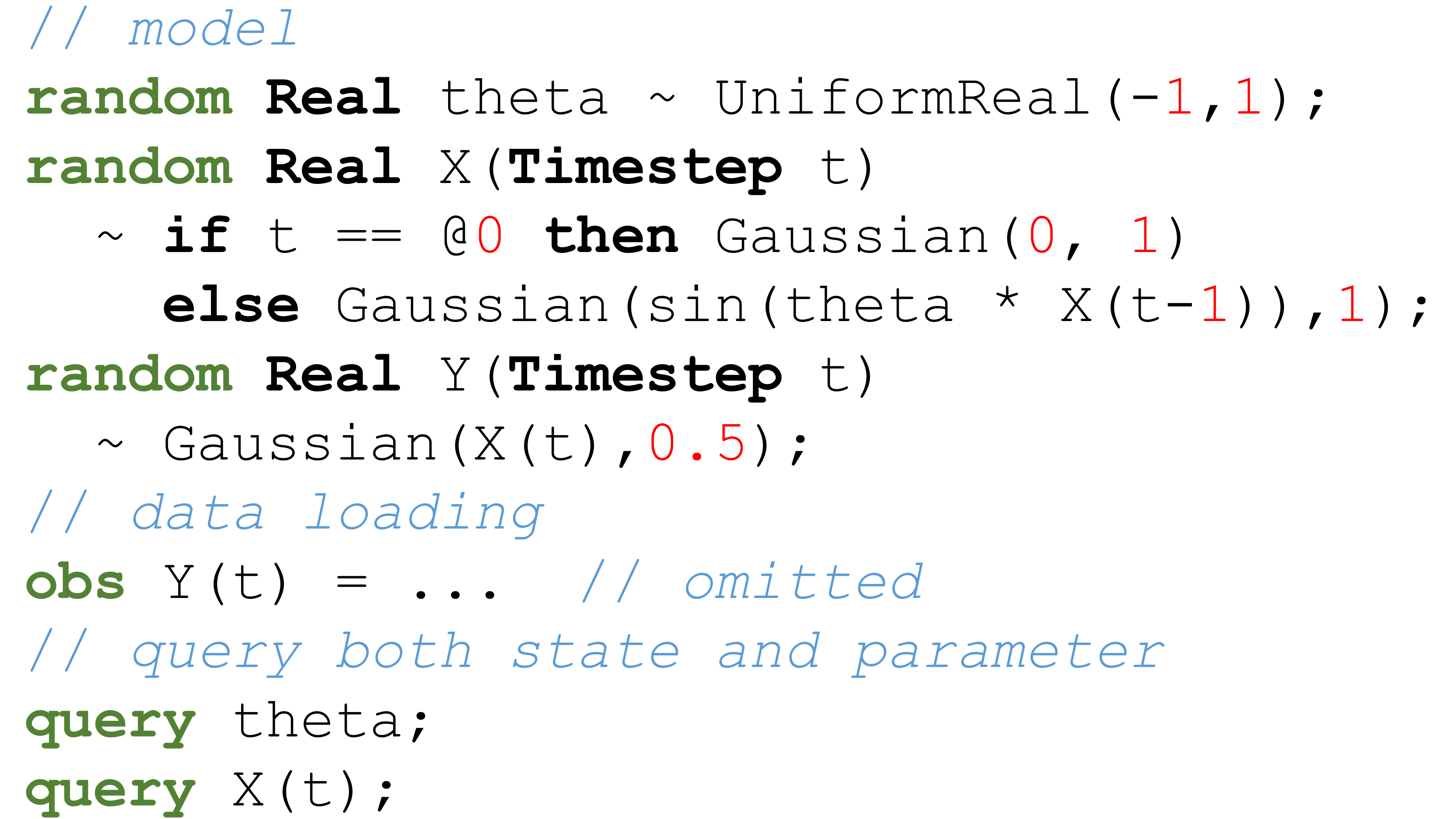}
%\end{minted}
%\end{scriptsize}
%\vspace{-3mm}
\caption{A simple dynamic model in \name: the SIN model considered in section \ref{sec:SIN_model}. \texttt{theta} is the parameter. \texttt{X(t)} are latent variables. \texttt{Y(t)} are observed.}\label{fig:sin-model}
\end{figure}

We have implemented \filter in an inference engine \name with a flexible probabilistic programming interface. \name utilizes the syntax of BLOG~\cite{milch2005blog}, a high-level modelling language to define probabilistic models (e.g., Figure~\ref{fig:sin-model}). \name analyzes the model and automatically generates model-specific inference code in C++. User can use \filter on any model written in \name. 
%\begin{minted}{c}
%int main() {
%return 0;
%}
%\end{minted}

\subsubsection{Memory Efficient Computation}

The particle filtering framework for \filter is often memory-intensive for applications with a large amount of data. Moreover, inefficient memory management in the inference engine can also result in tremendous overhead at runtime, which in turn hurts the estimation accuracy within a fixed amount of computational budget. Our black-box inference engine also integrates the following compilation optimizations for handling practical problems.

\textbf{Memory Pre-allocation: }
%Memory management is a critical issue for particle filter algorithms since it is often necessary to launch a large number of particles in practice.
%Offline systems (or offline algorithms) records all particles in the history which reduces the maximum number of possible particles. 
Systems that does not manage the memory well will repeatedly allocate memory at run time: for example, dynamically allocating memory for new particles and erase the memory belonging to the old ones after each iteration. This introduces significant overhead since memory allocation is extremely slow. In contrast, \name analyzes the input model and allocate the minimum static memory for computation: if the user specifies to run $N$ particles, and the Markov order of the input model is $D$, \name will allocate static memory for $(D+1)*N$ particles in the target code. When the a new iteration starts, we utilize a rotational array to re-use the memory of previous particles.

\hide{

\textbf{Lightweight Memoization: }
In Biips, the compiled target code will sample variables explicitly according to their declaration order in the model. It becomes feasible for Biips since any sorts of condition statements (i.e., if statement) are not supported, which results in a fixed dependency graph. However, the very restricted syntax of Biips is often infeasible for complicated real-world  applications. Notice that adopting an expressive language interface (i.e. syntax of BLOG) might lead to time-varying dependencies between random variables. \name memoizes the value for each random variables already sampled. An example compiled code fragment for the sin model (\ref{eq:sin_model}) is shown below.

\begin{scriptsize}
\begin{minted}{cpp}
class TParticle { public:
// value of x and y at current timestep
  double val_x, val_y;
// flag of whether x/y is sampled
  int mark_x, mark_y; 
}particles[N][DEP]; // particle objects
// getter function of y(t)
double get_y(int t) { 
// get the corresponding particle
  TParticle &part = get_particle(t); 
  // memoization for y(t)
  if(part.mark_y == cur_time) return part.val_y; 
  part.mark_y = cur_time;
// sample y(t), call getter function of x(t)
  part.val_y = sample_normal(
               sin(get_theta() * get_x(t-1)), 0.5);
  return part.val_x; }
\end{minted}
\vspace{-2mm}
\end{scriptsize}

This code fragment demonstrate the basic data structures as well as the memoization framework in the target code. \texttt{N} denotes the number of particles.  \texttt{DEP} is equal to the Markov order of the model plus 1.
In the memoization framework, \name allocates static memory for all the random variables within a particle data structure and generates a flag for each random variable denoting whether it has been sampled or not. Each random variable also has its \emph{getter} function. Whenever accessing to a random variable, we just call its getter function.
In practice, memoization causes negligible overhead.

Memoization also occurs in Figaro and BLOG. However, every random variable in Figaro and BLOG has a string ``name'', and a generic hashmap is used for value storage and retrieval, which brings a great deal of constant factor at run time. 

\textbf{Other Optimizations: } 

}

\name also avoids dynamic memory allocation as much as possible for intermediate computation step. 
For example, consider a multinomial distribution. When its parameters change, a straightforward implementation to update the parameters is to re-allocate a chunk of memory storing all the new parameters and pass in to the object of the multinomial distribution. In \name,  this dynamic memory allocation operation is also avoided by pre-allocating static memory to store the new parameters.

%\subsubsection{Computational Efficiency}
\textbf{Efficient Resampling: }
Resampling step is critical for particle filter algorithms since it requires a large number of data copying operations. Since a single particle might occupy a large amount of memory in real applications, directly copying data from the old particles to new ones induce substantial overhead.

In \name, every particle access to its data via an indirect pointer. As a result, redundant memory copying operations are avoided by only copying the pointers referring to the actual particle objects during resampling.
Note that each particle might need to store multiple pointers when dealing with models with Markov order larger than 1 (i.e., $x_k$ depends on both $x_{k-1}$ and $x_{k-2}$, which is supported in \name). 

\name also enhances program locality to speed up resampling.
%, when the number of particle becomes large, copying a large number of pointers is also expensive at run-time. To make the process even faster,
In the compiled code, the indexes of the array which stores the pointers are carefully aligned to take the advantage of memory locality when those pointers are copied. 

In our experiment, when dealing with small models where the resampling step takes a large fraction of overall running time,  \name achieves over 3x to 6x speedup against the fastest benchmark toolkit for PF and PMCMC.
%\reminder{Yi will correct 3x to 5-6x. Do we need to mention that this speed-up is for pf and pmcmc?}

\hide{
%An example compiled code is shown below.

\begin{scriptsize}
\begin{minted}{cpp}
// indirect pointers to the actual particle
TParticles* part_ptrs[N][DEP]; 
TParticle& get_particle(int t) {// rotational array
  return *part_ptr[cur_part][t % DEP]; 
  }
\end{minted}
\vspace{-1.5mm}
\end{scriptsize}

\textbf{Locality: }

Besides using pointer references to reduce the amount of moving data, The fragment of code used in resampling step is shown below.

\begin{scriptsize}
\begin{minted}{cpp}
void resample_ptr( int* target,
  TParticle* part_ptr[N][DEP], //storage of  pointers
  TParticle* backup_ptr[N][DEP]) // temporary storage 
{ for (int i = 0; i < N; ++i) {
    // pos: index of particle to copy data from
    int& pos = target_ptr[i];
    // move continuous range of data
    memcpy(backup_ptr[i], part_ptr[pos],
           sizeof(TParticle*)* DEP);  }
  std::memcpy(ptr_temp_memo, backup_ptr,
              sizeof(TParticle*)* DEP * N);}
\end{minted}
\vspace{-1.5mm}
\end{scriptsize}

In the preceding code fragment, the first dimension of the storage array, \texttt{part\underline{ }ptr}, corresponds to the particle index. While the second dimension corresponds to the current iteration. In this case, when copying the pointers during resampling step, all the memory are continuously located in the memory space, which greatly reduces the constant factor at run time.
}

\subsubsection{Algorithm Integration: }
Currently \name supports the bootstrap particle filter, Liu-West filter, PMCMC and \filter. User can specify any of these algorithms as the inference algorithm as well as the number of particles. The Markov order of the input model will be automatically analyzed.

When choosing \filter, \name analyzes all the static parameters in the model and compiles different approximation distributions for different types of random variables. At the current stage, \name supports Gaussian and Mixtures of Gaussian to approximate continuous variables and multinomial distribution for discrete variables. More approximation distributions are under development. For sampling approach, by default \name assumes all the parameters are independent and uses deterministic sampling when possible. User could also ask \name to generate random samples instead. Furthermore, the number of approximation samples ($M$), the number of mixtures ($L$) can also be specified. By default, $M=7$ and $K=10$.

\subsection{Complexity}
The time and space complexity of \filter is $O(NMT)$ over $T$ time steps for generating $N$ particles for $\theta$ and $x$ as well as extra $M$ samples for each particle path to update the sufficient statistics through the moment matching integrals. Setting $N$ and $M$ adequately is crucial for performance. Small $N$ prevents \filter exploring the state space sufficiently whereas small $M$ leads to inaccurate sufficient statistics updates which will in turn result in inaccurate parameter estimation.

Note that when taking $M$ MCMC iterations, the PMCMC algorithm also has complexity of $O(NMT)$ for $N$ particles over $T$ time steps. However, PMCMC typically requires a large amount of MCMC iterations for mixing properly while very small $M$ is sufficient for \filter to produce accurate parameter estimation.

Moreover, the actual running time for \filter is often much smaller than its theoretical upper bound $O(NMT)$. Notice that the approximation computation in \filter only requires the local data in a single particle and approximation results does not influence the weight of that particle. Hence, one important optimization specialized for \filter is that \name resamples all the particles prior to the approximation step at each iteration and only updates the approximation distribution for those particles that do not disappear after the resampling step. Notice that it is often the case that a small fraction of particles have significantly large weights. Hence in practice, as shown in the experiment section, the overhead due to the $M$ extra samples only causes negligible overhead comparing with the plain particle filter. In contrast, the theoretical time complexity is tight for PMCMC.
% We investigate the dependency of \filter on $N$ and $M$ in the experiment section.

%\subsection{Black Box Inference (SPEC)}

%\section{The Black-Box Inference Engine (SPEC)}\label{sec:system}
%\input{04system.tex}

\section{Experiments}\label{sec:experiment}
%We compare \name with BLOG,\footnote{ \url{http://bayesianlogic.github.io}}, Figaro\footnote{\url{https://www.cra.com/work/case-studies/figaro}} and Biips\footnote{R-Biips v0.10.0, \url{https://alea.bordeaux.inria.fr/biips/}}.
The experiments were run on three benchmark models\footnote{Code can be found in the supplementary material.}:
\begin{inparaenum}
\item \sinmodel: a nonlinear dynamical model with a single continuous parameter; \item \slammodel: a simultaneous localization and Bayesian map learning problem with 20 discrete parameters; 
\item \birdmodel: a 4-parameter model to track migrating birds with real world data.
\end{inparaenum}
We compare the estimation accuracy of \filter against Liu-West filter and PMCMC within our compiled inference \name. We also compare our \name system with other state-of-the-art toolboxes, including Biips\footnote{R-Biips v0.10.0, \url{https://alea.bordeaux.inria.fr/biips/}} and LibBi\footnote{LibBi 1.2.0. \url{http://libbi.org/}}. Since Biips and LibBi do not support discrete variable or conditional-dependency in the model, we are only able to compare against them on the \sinmodel model.

The experimental machine is a computer equipped with Intel Core i7-3520 and 16G memory. 
%All the inference systems run under single-thread mode. For random number generator, \name uses standard C++ \verb|<random>| library.

\subsection{Toy nonlinear model}
\label{sec:SIN_model}
%We consider the sin model in equation (\ref{eq:sin_model})
We are considering the following nonlinear model (the modeling code is in Figure~\ref{fig:sin-model}):% in \cite{erol2013extended}.
\beq
	x_t &=& \sin(\theta x_{t-1}) + v_t, \quad v_t \sim \gauss(0,1) \notag  \\
	y_t &=&x_t +w_t, \quad w_t \sim \gauss(0,0.5^2)
\label{eq:sin_model}
\eeq
where $\theta^{\star}=-0.5$ and $\theta \sim \gauss(0,1)$ and $x_0\sim \gauss(0,1)$. We generate $5000$ data points using $\theta^\star$ from this model. Notice that it is not possible to use Storvik filter \cite{storvik2002particle} or the particle learning \cite{lopes2010particle} algorithm for this model as sufficient statistics do not exist for $\theta$.

%\reminder{We could remove the following para}
%As a sanity check, we first show that for a given $x_{0:t}$, our approximation scheme is capable of approximating $p(\theta \mid x_{0:t})$. For the Gaussian approximation, we use the Unscented transform and the Gauss-Hermite quadrature with different number of quadrature points. The KL-divergence between the posterior of interest and our approximation is illustrated in Figure \ref{fig:kl_div}.
%\begin{figure}[h]
%\centering
%	\includegraphics[width=2in]{fig/KLdiv_T5000.eps}
%\caption{KL divergence between true posterior and $q_t(\theta)$ for different quadrature rules}
%\label{fig:kl_div}
%\end{figure}
%
%\vspace{-1em}
%It can be seen that, the KL-divergence tends to decrease with more data points. This intuition agrees with the Gaussian ansatz analysis in \cite{opper1998bayesian}. Furthermore, it should be noted that increasing the number of quadrature points ($M$) leads to better approximations. We have noticed that increasing $M$ more than 7 doesn't improve the approximation accuracy for this model.  

\hide{
\textbf{Efficiency: }

We run benchmark PPLs using the basic particle filter algorithm (PF) with different number of particles. For \name, we run both PF and \filter. \filter adopts Gaussian approximation with $M=7$ Gauss-Hermite quadrature points. 

The results in Fig.~\ref{fig:sin_speed} show that \name achieves two orders of magnitudes speedup than benchmark systems. Specifically for 5000 particles, \name with PF uses 2.21s, which is over 260x faster than other PPLs, while \filter with $M=7$ uses 8.4365s, which is over 70x faster. Noticed that although \filter generates extra 7x samples for each iteration, the overall running time is merely 3x slower than PF. %The overhead is negligible compared to other systems.

\textbf{Accuracy:}

}
We evaluate the mean square error of the estimation results over 10 trials within a fixed amount of computation time given our generated data points for Liu-West filter, PMCMC and \filter. Note that all these algorithms are producing samples while ground truth is a point estimate. We take the mean of the samples for $\theta$ produced by Liu-West and \filter at the last time step. For PMCMC with $M$-MCMC iterations, we take the mean of the last $M/2$ samples and leave the first half as burn-in.

We choose the default setting for \filter: Gaussian approximation with $M$ samples. For PMCMC, we experiment on different number of particles (denoted by $N$). For proposal of $\theta$, we use a local proposal of truncated Gaussian distribution. We also perform a grid search over the variance of the proposal and only report the best one.

\begin{figure}[tb]
\centering
\includegraphics[width=0.48\textwidth]{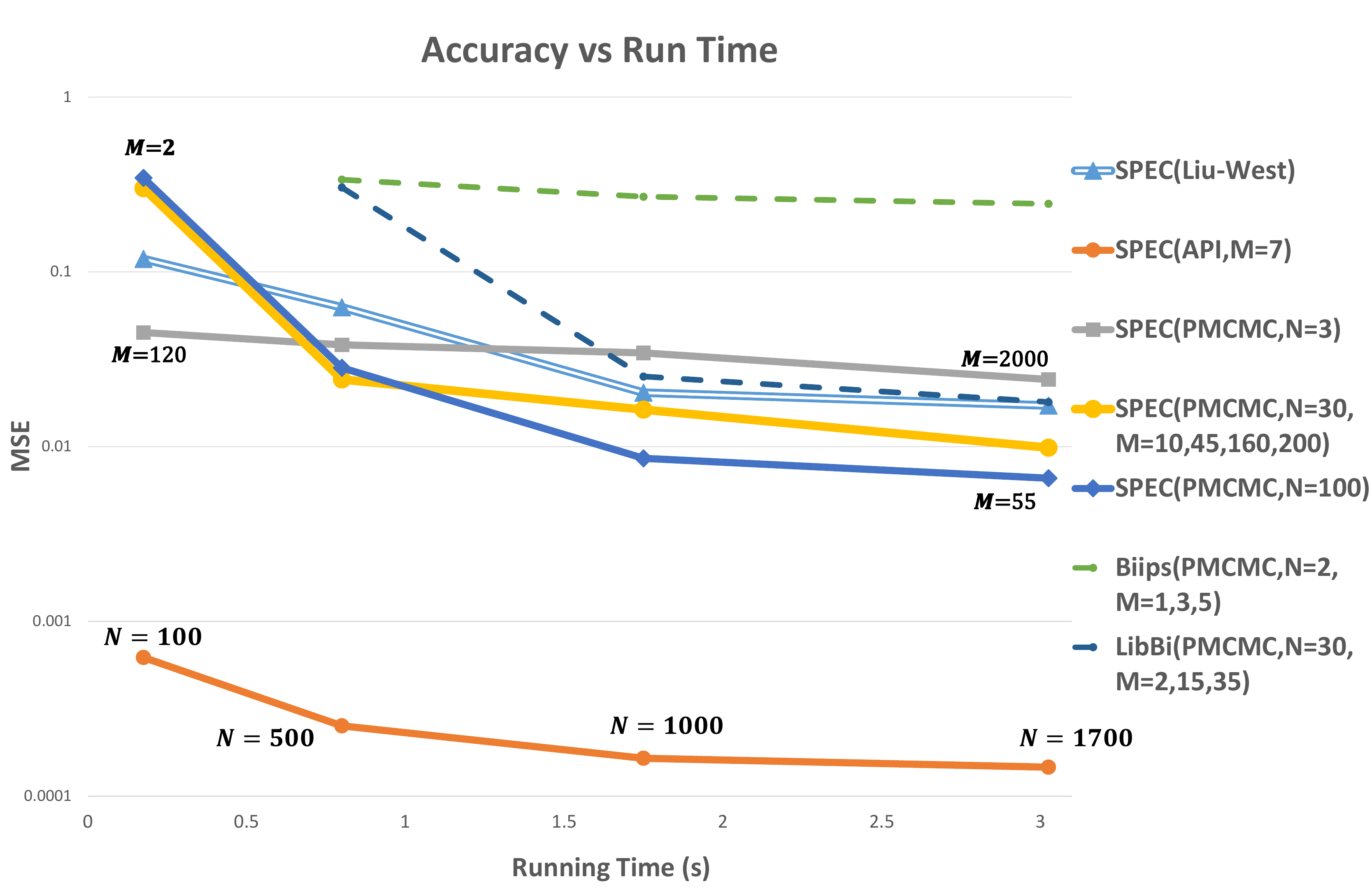}
\vspace{-1.5em}
\caption{The estimation accuracy of different algorithms with various parameters and implementations on the \sinmodel model. \filter produces order-of-magnitude more accurate estimation than Liu-West and PMCMC. The PMCMC by \name is 6x faster than LibBi.}\label{fig:sin-plot}
\end{figure}

In order to investigate the efficiency of \name, we also compare the running time of PMCMC implementation of \name against Biips and LibBi. The results are shown in Figure~\ref{fig:sin-plot}.

\filter produced a result of two orders of magnitude smaller error within the same amount of running time: it quickly produces accurate estimation (error: $1.6*10^{-4}$) using only 1000 particles and 1.5 seconds, which is still 50x smaller than the best PMCMC result after 3 seconds. For PMCMC in LibBi, we utilize $N=30$ particles. Within 3 seconds, it only produces 35 MCMC samples while \name finishes 200 iterations.
For PMCMC in Biips, just one MCMC step with one particle takes 0.6s, within which \name could already produce high quality estimation.

\textbf{Parallel Particle Filter in LibBi: }
LibBi supports advanced optimization choices, including vectorization (SSE compiler option), multi-thread (openmp) and GPU (cuda) versions. We experimented on all these advanced versions and chose the fastest one in Figure~\ref{fig:sin-plot}: the single-thread with SSE compiler option.  We also experiment on 100, 1000 and 10000 particles on the \sinmodel model in LibBi's with multi-thread and GPU option. The parallel versions are still slower than the single thread one on 100 and 1000 particles. For 10000 particles, GPU and multi-thread versions are twice faster than the single-thread version. Note that even with 10000 particles, the inference code generated by \name is still 20\% faster than the parallel versions by LibBi. In practice, parallelization often incurs significant communication and memory overhead, especially on GPU. Also, due to the resampling step, it is non-trivial to come up with an efficient parallel compilation approach for particle filtering. This experiment demonstrates the importance of memory management in practical setting: with memory efficient computations, even a sequential implementation can be much faster than the parallel version.
%We also test 200 MCMC steps with 100 particles in Biips, which takes more than 1 hour to finish. The result is still of inferior quality.

\textbf{Bimodal Variant: }
The above \sinmodel defines a unimodal posterior on the parameter. We are slightly modifying the model as follows in order to explore our algorithm's performance on multimodal posteriors.
\beq
	x_t &=& \sin(\theta^2 x_{t-1}) + v_t, v_t \sim \gauss(0,1) \nn  \\
	y_t &=&x_t +w_t, w_t \sim \gauss(0,0.5^2)
\eeq
Due to the $\theta^2$ term, $p(\theta \mid y_{0:t})$ will be bimodal. We execute \filter with $10^3$ particles and $M=7$ using $K$ mixtures of Gaussian for approximation. We also ran the PMCMC with 100 particles using truncated Gaussian proposal. We ran PMCMC for 10 minutes to ensure mixing.
%For ensure convergence, we ranThe slowest one finished in 6 seconds. 
We measure the performance for \filter with $K=2$, $K=5$ , $K=10$ as well as PMCMC. The histograms of the samples for $\theta$ are illustrated in Figure~\ref{fig:hist_bimodal}. Comparing with the result by PMCMC, \filter indeed approximates the posterior well. Note that even with $10^5$ particles, Liu-West filter cannot produce a bimodal posterior.

For $K=2$, \filter was only able to find a single mode. For both $K=5$ and $K=10$, \filter successfully finds two modes in the posterior distribution, though the weights are more accurate in the case of $K=10$ than $K=5$. This implies that increasing the number of mixtures used for approximation helps improving the probability of finding different modes in the posterior distribution. 

\begin{figure}[tb]
\centering
\includegraphics[width=0.4\textwidth]{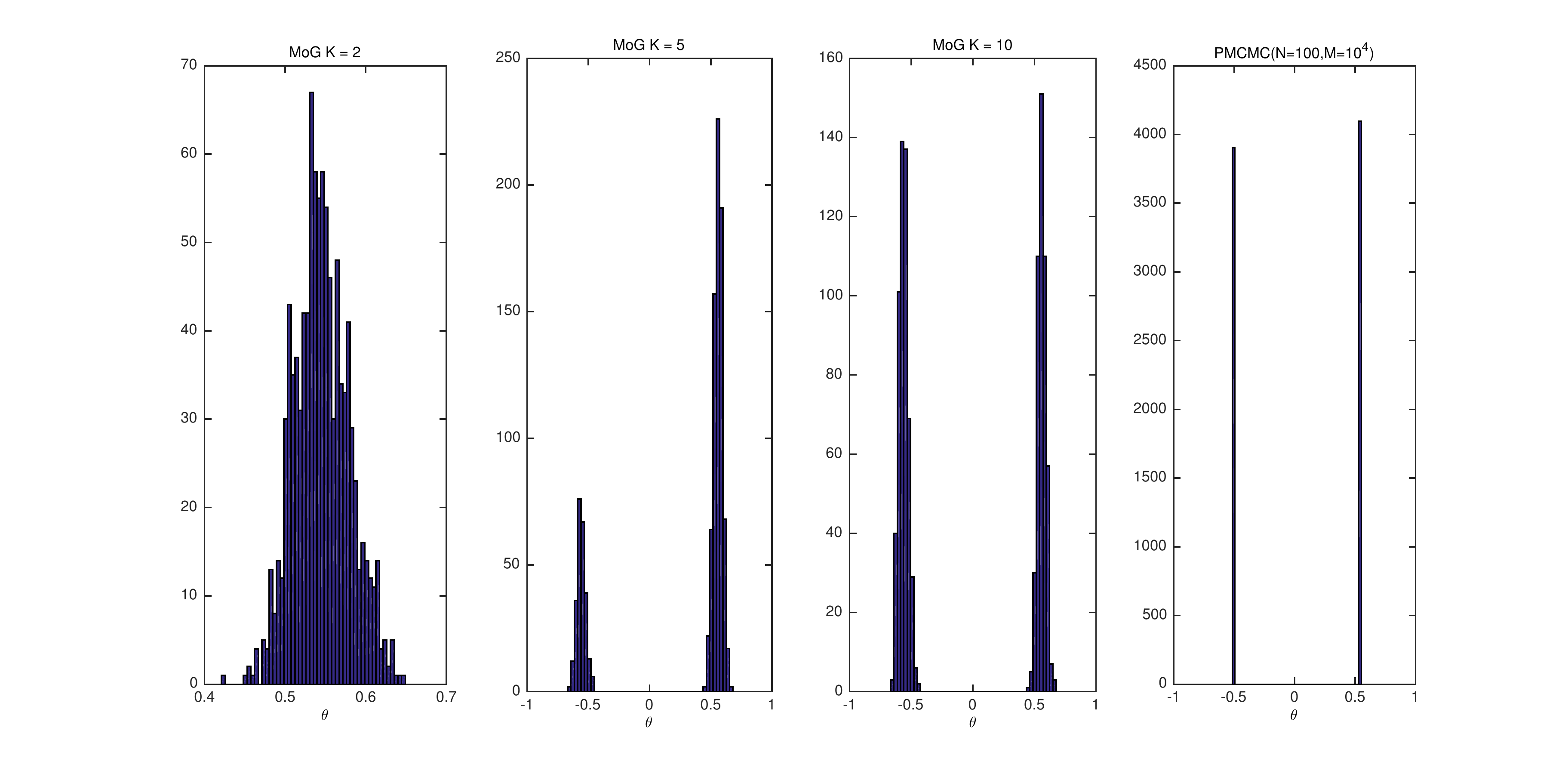}
\vspace{-1em}
\caption{The histograms of the produced samples for $\theta$ in the multimodel example by \filter with different parameters and by PMCMC. \filter indeed approximates the true posterior well.}\label{fig:hist_bimodal}
\end{figure}

\hide{
\begin{figure*}[tb]
\centering
\subfigure[Efficiency on \sinmodel]{
   \includegraphics[width=0.23\textwidth]{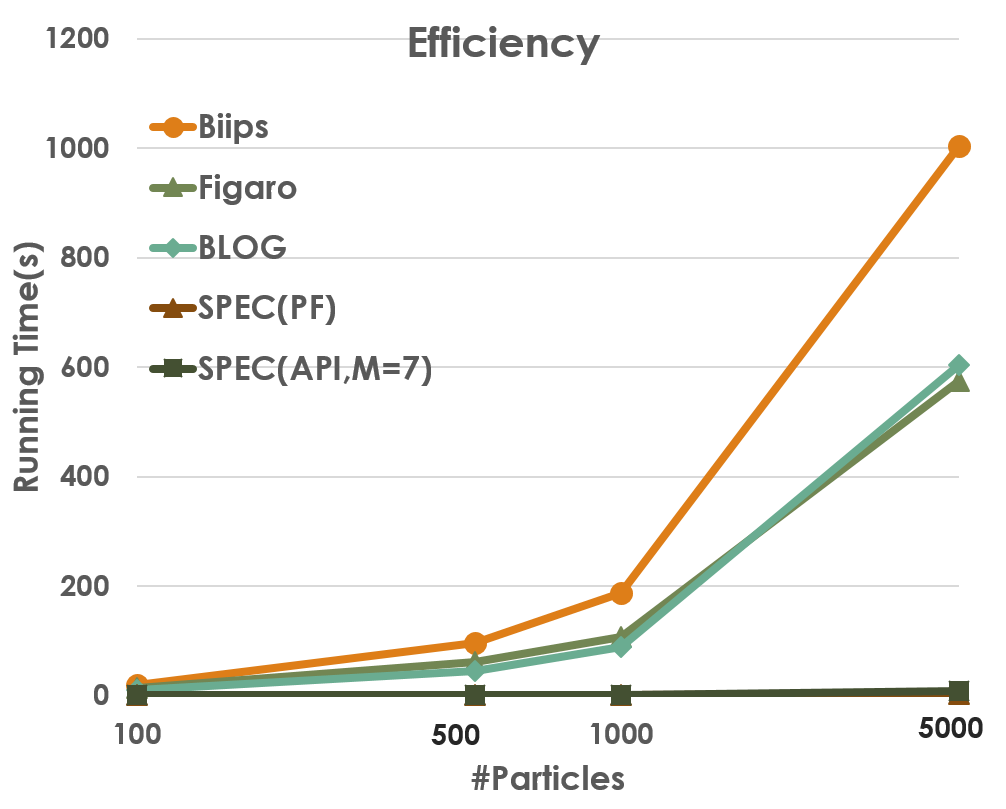}
   \label{fig:sin_speed}
}
\subfigure[Accuracy on \sinmodel]{
   \includegraphics[width=0.23\textwidth]{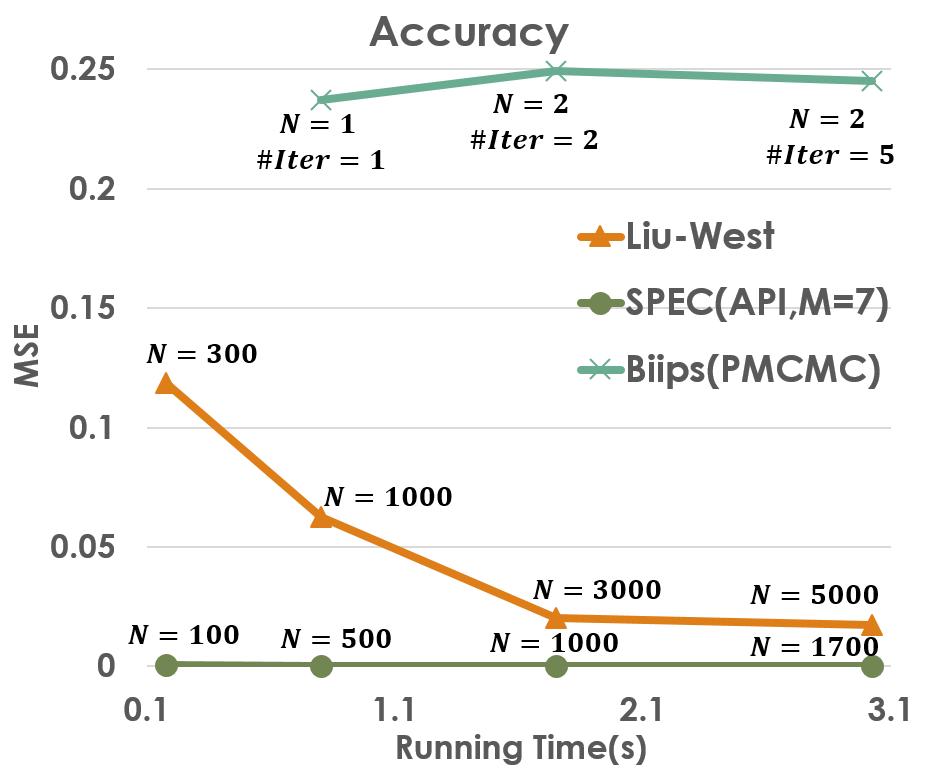}
  \label{fig:sin_err}
}
\subfigure[Efficiency on \slammodel]{
   \includegraphics[width=0.23\textwidth]{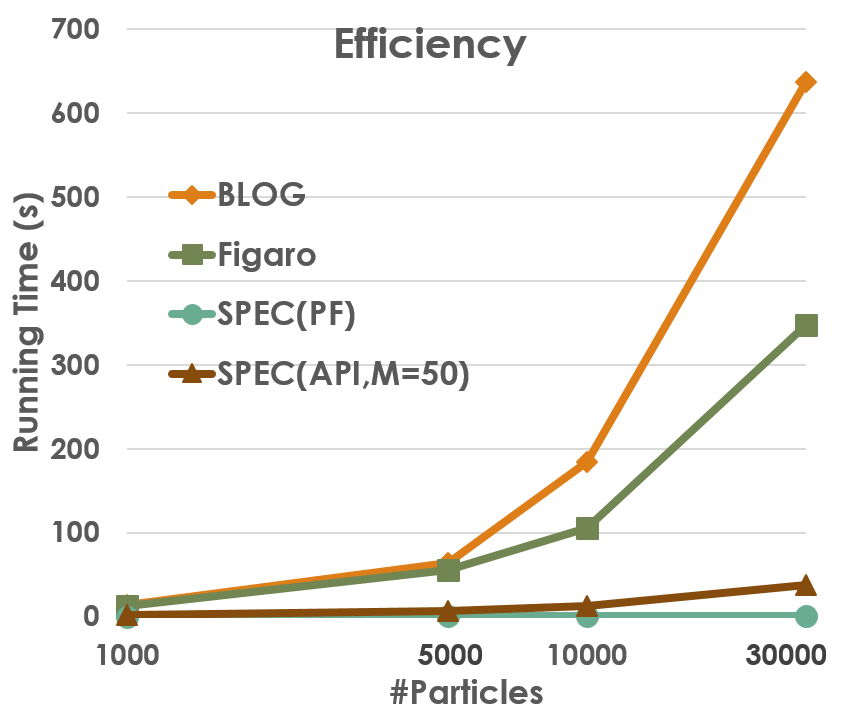}
 \label{fig:slam_speed}
}
\subfigure[Accuracy on \slammodel]{
   \includegraphics[width=0.23\textwidth]{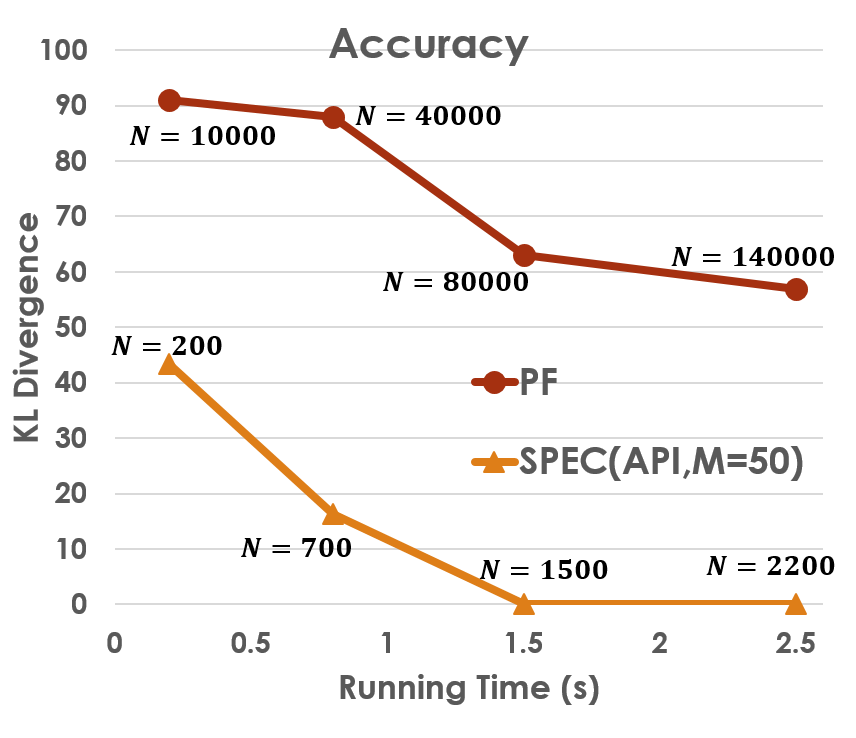}
   \label{fig:slam_err}
}
\subfigure[Efficiency on \birdmodel]{
   \includegraphics[width=0.23\textwidth]{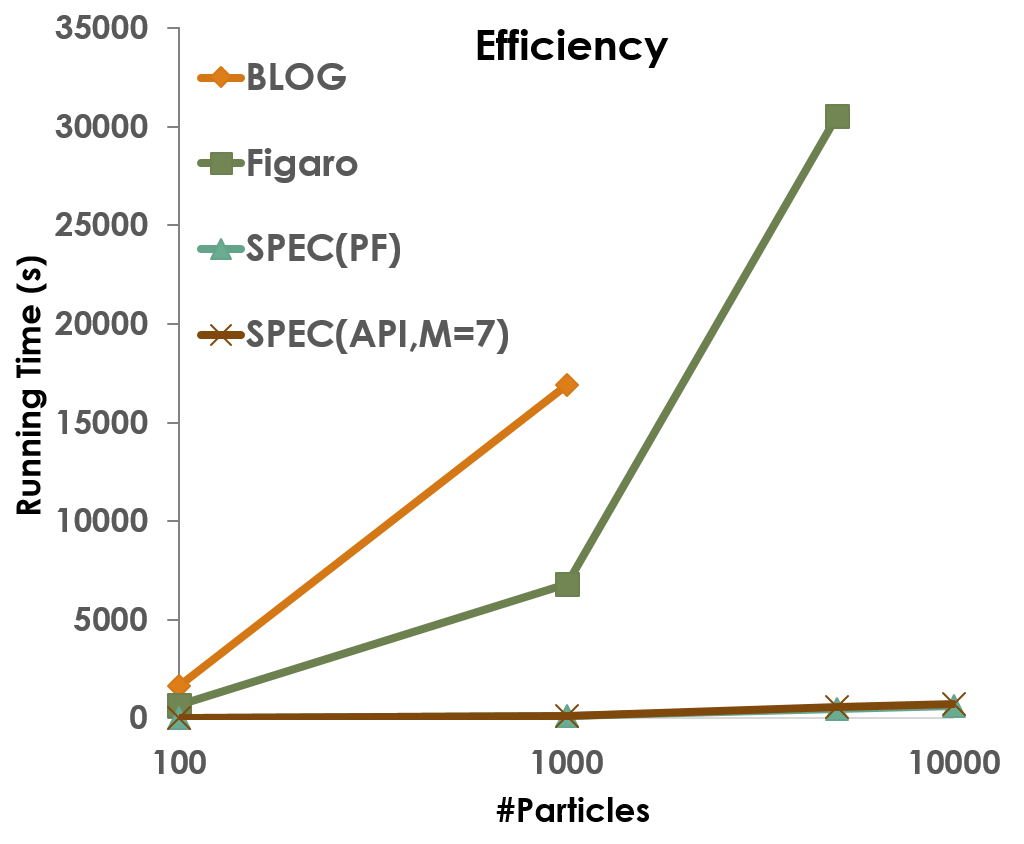}
   \label{fig:bird_speed}
}
\subfigure[Accuracy on \birdmodel]{
   \includegraphics[width=0.23\textwidth]{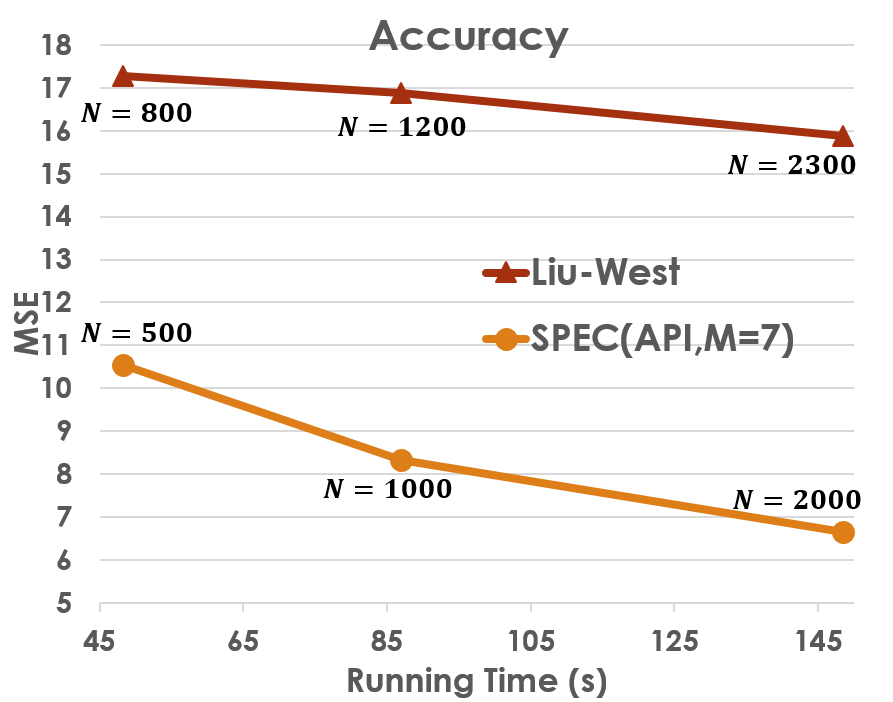}
   \label{fig:bird_err}
}
\subfigure[Histograms of $p(\theta|y_{0:T})$ in \sinmodel by \filter with $K=2,5,10$]{
   \includegraphics[width=0.23\textwidth]{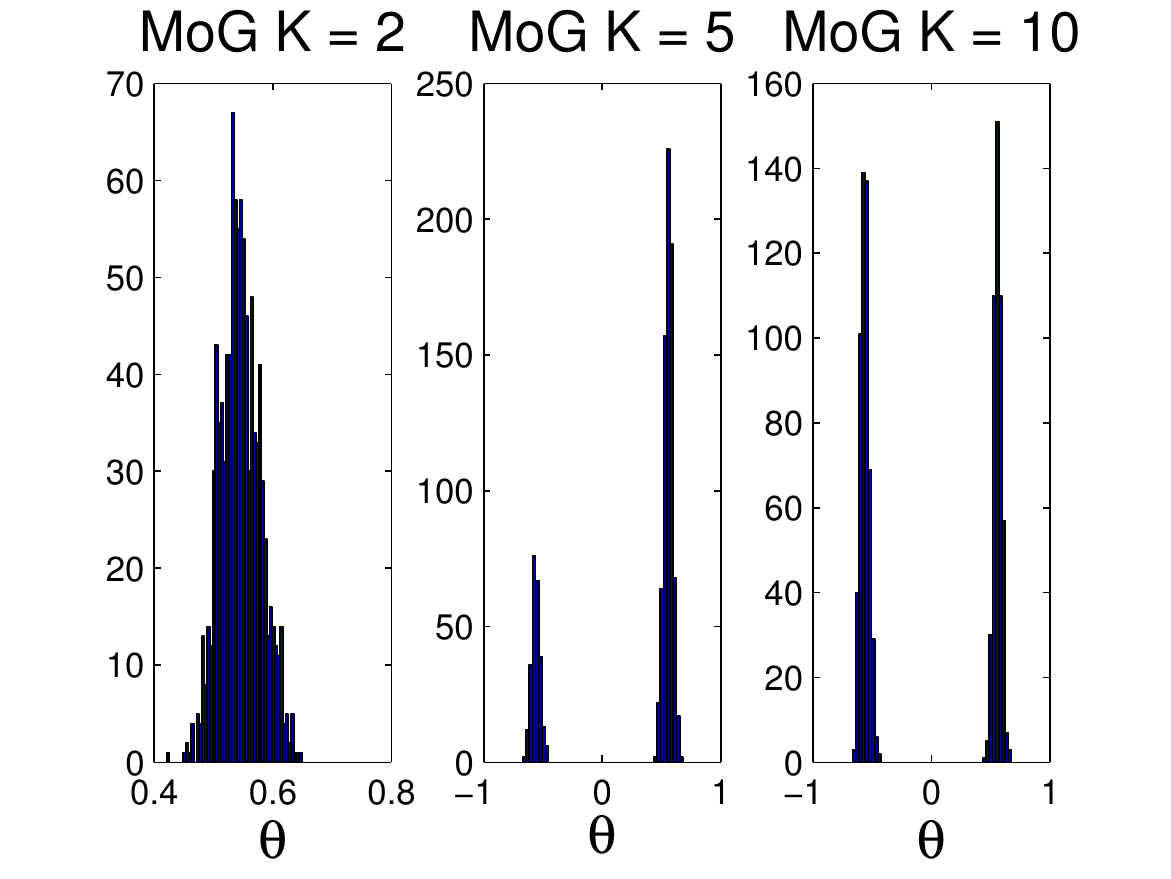}
   \label{fig:hist_bimodal}
}
\subfigure[\filter with different settings on \slammodel]{
   \includegraphics[width=0.20\textwidth]{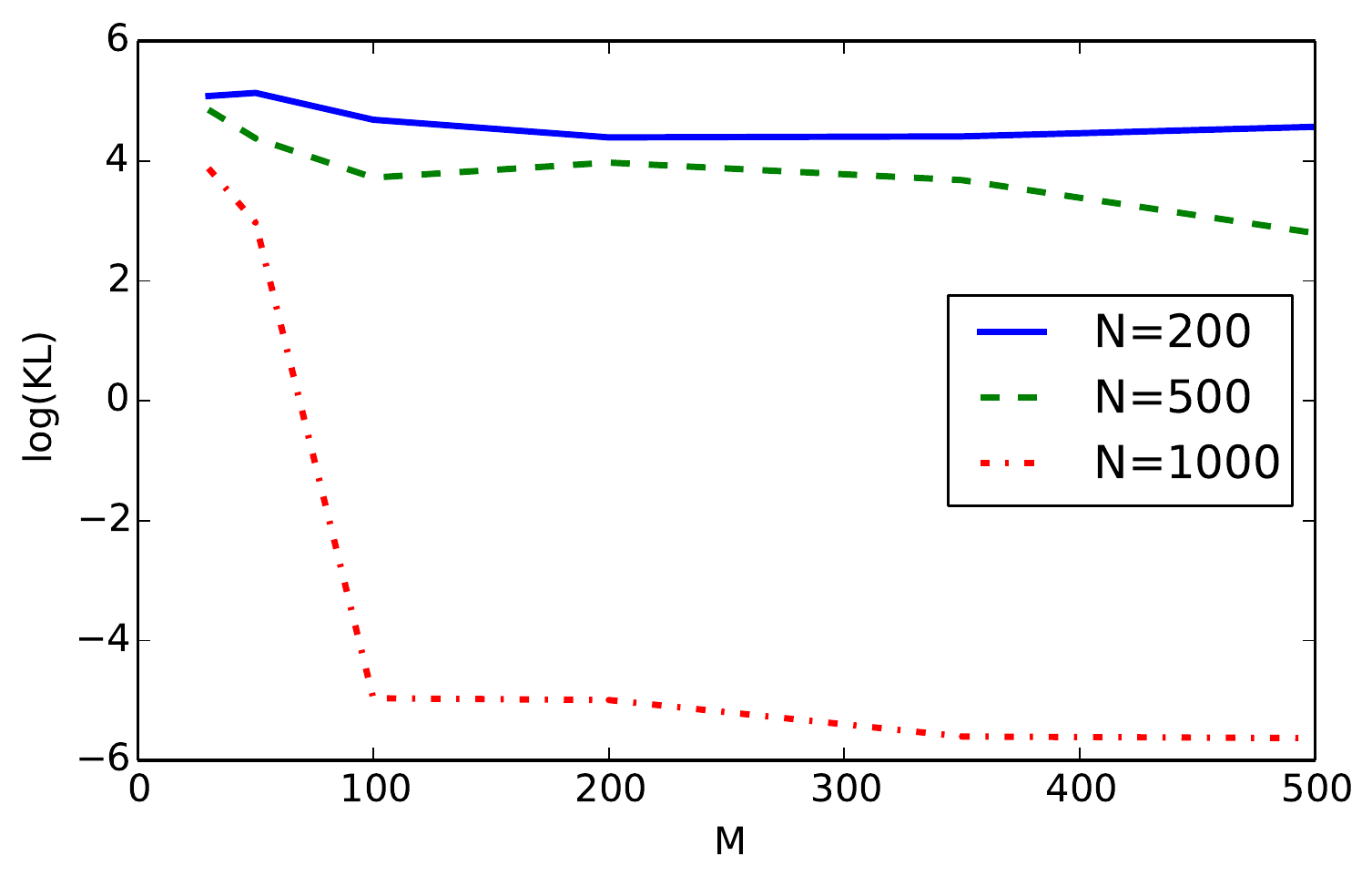}
 \label{fig:slam_param}
}
\vspace{-1em}
\caption{Performance Plots in \sinmodel, \slammodel, \birdmodel Experiments.}
\vspace{-1em}
\end{figure*}
}
% For the bimodal posterior, both methods perform comparably. 
% %We have also noticed that, in this case any $K$ greater than or equal to 2 works equally well.
% The mixture components either move towards the modes or they get down-weighted as expected. 

% The biggest benefit of \filter is that there is no need for a hand-crafted approximation. EPF requires the \emph{separable form} \cite{erol2013extended}, defined by the Taylor approximation to be inputted by the user. Note that, coming up with proper polynomial approximations for high-dimensional complicated models (i.e. the bird migration model) is not an easy task. Furthermore, at each time step for each particle, EPF needs to sample from a log-polynomial density and there are no black box samplers for such densities (in this experiment, we used an MCMC sampler). \filter samples from simple distributions and does not require any hand-crafted approximation.

\subsection{Simultaneous localization and mapping}
% \begin{figure*}[tb]
% \vspace{-0.1in}
% \centering
% \subfigure[Exact]{
%     \includegraphics[width=1.5in]{fig/exact_map.eps}
%     \label{fig:exact_map}
% }
% \subfigure[RBPF] {
%     \includegraphics[width=1.5in]{fig/rbpf_map.eps}
%     \label{fig:rbpf_map}
% }
% \subfigure[Proposed]{
%   \includegraphics[width=1.5in]{fig/proposed_map.eps}
%   \label{fig:proposed_map}
% }
% \subfigure[KL-divergence wrt $N$ \& $M$]{
%   \includegraphics[width=1.5in]{fig/kl_wrt_N_M.eps}
%   \label{fig:kl_map}
% }
% \vspace{-0.1in}
% \caption{
% \subref{fig:exact_map}: Exact posterior computed by the HMM filter.
% \subref{fig:rbpf_map}: Posterior computed by the RBPF with $N=1000$ particles.
% \subref{fig:proposed_map}: Posterior computed by the proposed approach with $N=1000$ and $M=50$ particles.
% \subref{fig:kl_map} KL-divergence between the exact posterior and the posterior computed by our approach with respect to $N$ and $M$ at time step $t=T$.
%  }
% \end{figure*}

We are considering a Simultaneous localization and mapping example (\slammodel) modified from \cite{bayesian2000murphy}. The map is defined as a 1-dimensional grid, where each cell has a static label (parameter to be estimated) which will be observed by the robot. More formally, the map is a vector of discrete random variables $M(i) \in \left\{1, \dots, N_{O} \right\}$, where $1 \leq i \leq N_L$. Neither the map nor the robot's location $L_t \in \left\{1, \dots, N_L \right\}$ is observed. The existing observations are the label of the cell at robot's current location and the action chosen by the robot. 

Given the action (move right ($\rightarrow$) or left ($\leftarrow$)) the robot moves in the direction of action with a probability of $p_a$ and stays at its current location with a probability of $1-p_a$ (i.e. robot's wheels slip). The prior for the map is a product of individual cell priors, which are all uniform. The robot observes the label of the cell, it is currently located in correctly, with a probability of $p_o$ and incorrectly with a probability of $1-p_o$.

In the original example, $N_L=8$, $p_a=0.8$, $p_o=0.9$, and 16 actions were taken. In our experiment, we enhance the model by setting $N_L=20$ and duplicating the actions in \cite{bayesian2000murphy} several times to finally derive a sequence of 164 actions. 

\hide{
\textbf{Efficiency: } We compare the running time of \name against BLOG and Figaro. Biips fails to compile this model due to its restricted syntax. For \name, we run PF as well as \filter with $M=50$ approximation samples. We use a fully factorized (Bernoulli) discrete distribution for approximation.

The results in Fig.~\ref{fig:slam_speed}, PF uses 0.596s for $3*10^4$ particles, which is over 500x speedup than BLOG and Figaro. For \filter with $M=50$, although theoretically it would be $N_L M=1000$ times slower than PF, it uses 37.296s for $3*10^4$ particles, which is merely 60x slower than PF and still 10x faster than Figaro (347.89s) and BLOG (637.75s).

\textbf{Accuracy: } }
We compare the estimation accuracy over particle filter (PF), PMCMC and \filter within \name. Notice that the Liu-West filter is not applicable in this scenario as artificial dynamics approach can only be performed for continuous parameters. For PMCMC, at each iteration, we only resample a single parameter using an unbiased Bernoulli distribution as the proposal distribution. For \filter we use $M=50$ approximation samples and a fully factorized (Bernoulli) discrete distribution for approximation.

Since it is still possible to compute the exact posterior distribution,  we run these algorithms within various time limits and measure the KL-divergence between the estimated distributions and the exact posterior.  The results in  Figure ~\ref{fig:slam_err} show that PF  drastically suffers from the sample impoverishment problem; PMCMC fails to get rid of a bad local mode and suffers from poor mixing rates while \filter successfully approximates the true posterior even with only 1500 particles.

Note that we also measure the running time of \filter against the plain particle filter on this model.  PF uses 0.596s for $3*10^4$ particles. For \filter with $M=50$, although theoretically it would be $N_L M=1000$ times slower than PF, it uses only 37.296s for $3*10^4$ particles, which is merely 60x slower than PF.

\begin{figure}[tb]
\centering
\includegraphics[width=0.48\textwidth]{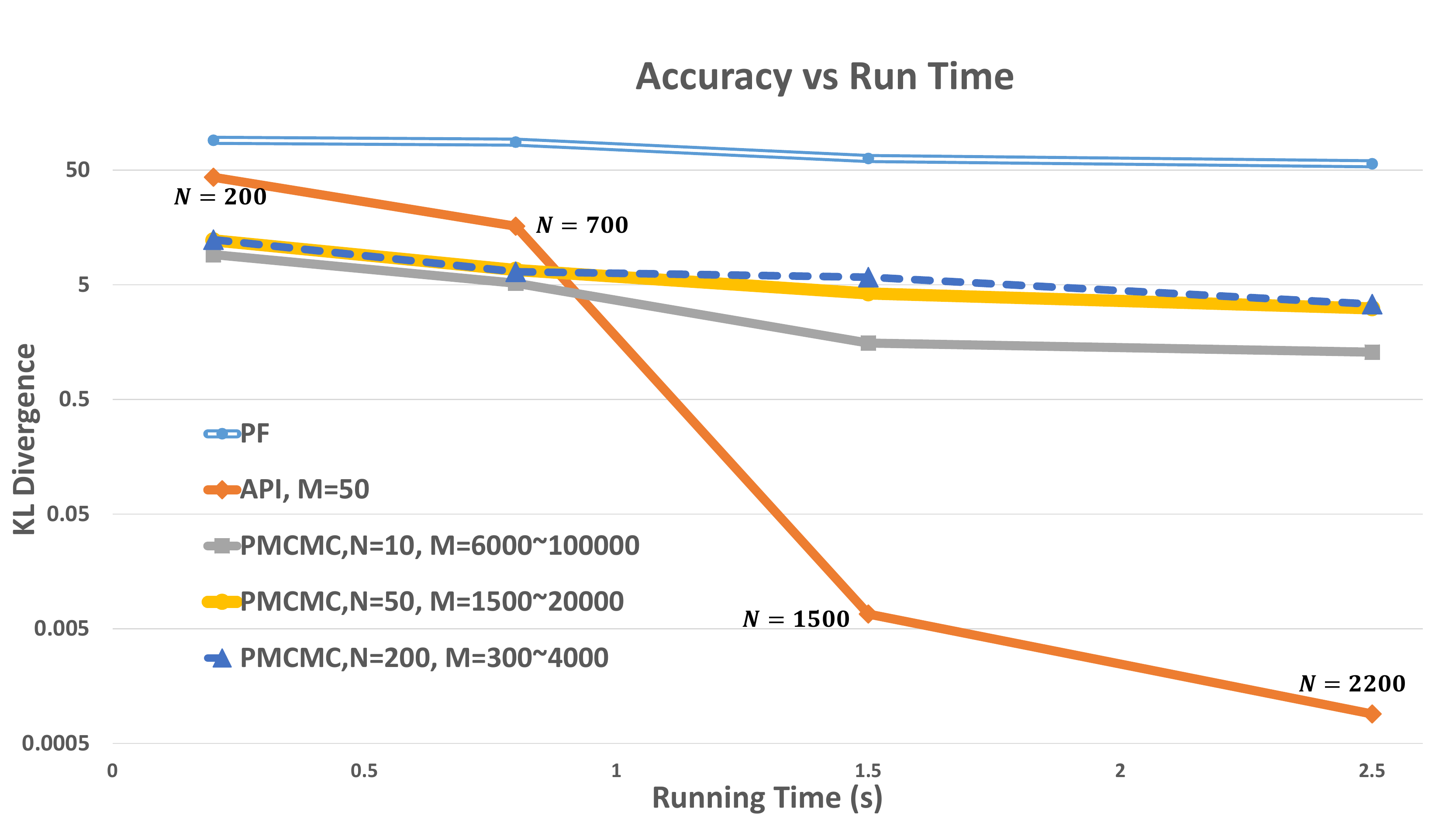}
\caption{The KL-Divergence between the true posterior and the estimation results produced by different algorithms.  PMCMC was trapped in a local model while \filter quickly produces order-of-magnitude more accurate estimation with only 1500 particles.}\label{fig:slam_err}
\end{figure}

\textbf{Choices of Parameters: } We experiment \filter with different parameters (number of particles $N$ and number of samples $M$) and evaluate the average log KL-divergence over 20 trials. The results in Figure~\ref{fig:slam_param} agree with theory. As $N$ increases the KL-divergence constantly decreases whereas after a certain point, not much gain is obtained by increasing $M$. This is due to the fact that for big enough $M$, the moment matching integrals are more or less exactly computed and the error is not due to the Monte Carlo sum but due to the error induced by the ADF projection step. This projection induced error cannot be avoided.

\begin{figure}[tb]
\centering
\includegraphics[width=0.35\textwidth]{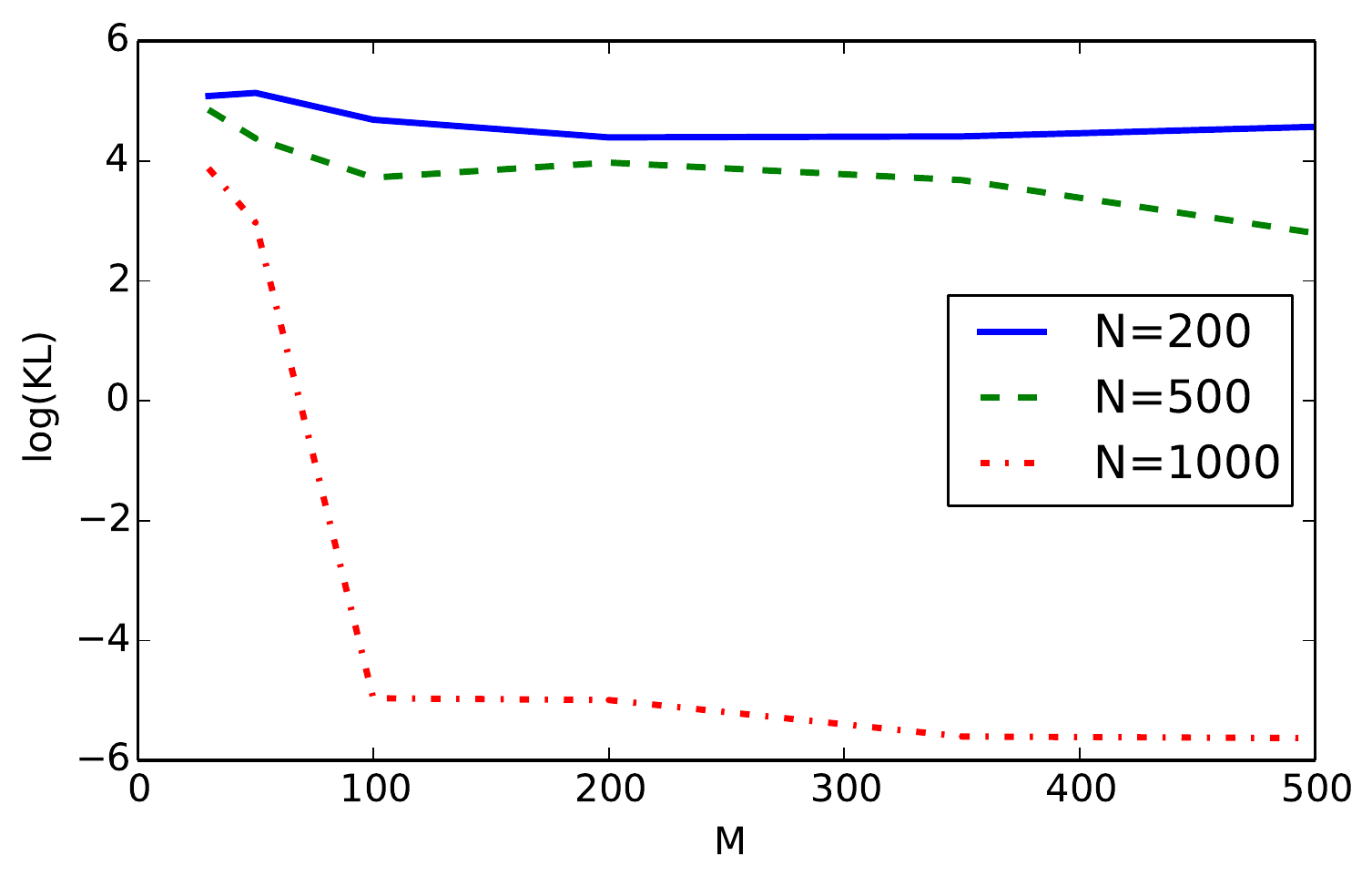}
\vspace{-1.5em}
\caption{\filter with various parameters on the \slammodel model.}\label{fig:slam_param}
\end{figure}

\subsection{Tracking bird migration}
The bird migration problem (\birdmodel) is originally investigated in \cite{elmohamed2007collective}, which
proposes a hidden Markov model to infer bird migration paths from a
large database of observations. We apply the particle
filtering framework to the bird migration model using the dataset officially released\footnote{\url{http://ppaml.galois.com/wiki/wiki/CP2BirdMigration}}. 
In the dataset, the eastern continent of U.S.A is partitioned into a 10x10 grid. There are roughly $10^6$ birds totally observed in the dataset. For each grid,
the total number of birds is observed over 60 days within 3 years. We
aim to infer the number of birds migrating at different grid locations
between two consecutive days.
To sum up, in the \birdmodel model, there are 4 continuous parameters with 60 dynamic states where each time step contains 100 observed variables and $10^4$ hidden variables.

\hide{
\textbf{Efficiency: }
We compare the running time of \name with PF and \filter, Figaro and BLOG on this model. Biips fails to express this complicated model. 

The results are shown in Fig.~\ref{fig:bird_speed} show. BLOG and Figaro do not produce an answer when the number of particles becomes large due to running out of the 16G memory. In this real application, \name with PF uses 104.136s and \filter uses 133.247s for 1000 particles, which are over 100x faster than BLOG ($1.7*10^4$s) and over 50x than Figaro ($6812$s). Note that in this real application,  \filter only results in 20\% overhead compared with PF.

\textbf{Accuracy: }
}

We measure the mean squared estimation error over 10 trials for \filter with Gaussian approximation ($M=7$), the Liu-West filter and PMCMC with truncated Gaussian proposal distribution within different time limits. The results are shown in the right part of Figure~\ref{fig:bird_err} again show that our \filter achieves much better convergence within a much tighter computational budget.

\begin{figure}[tb]
\centering
\includegraphics[width=0.48\textwidth]{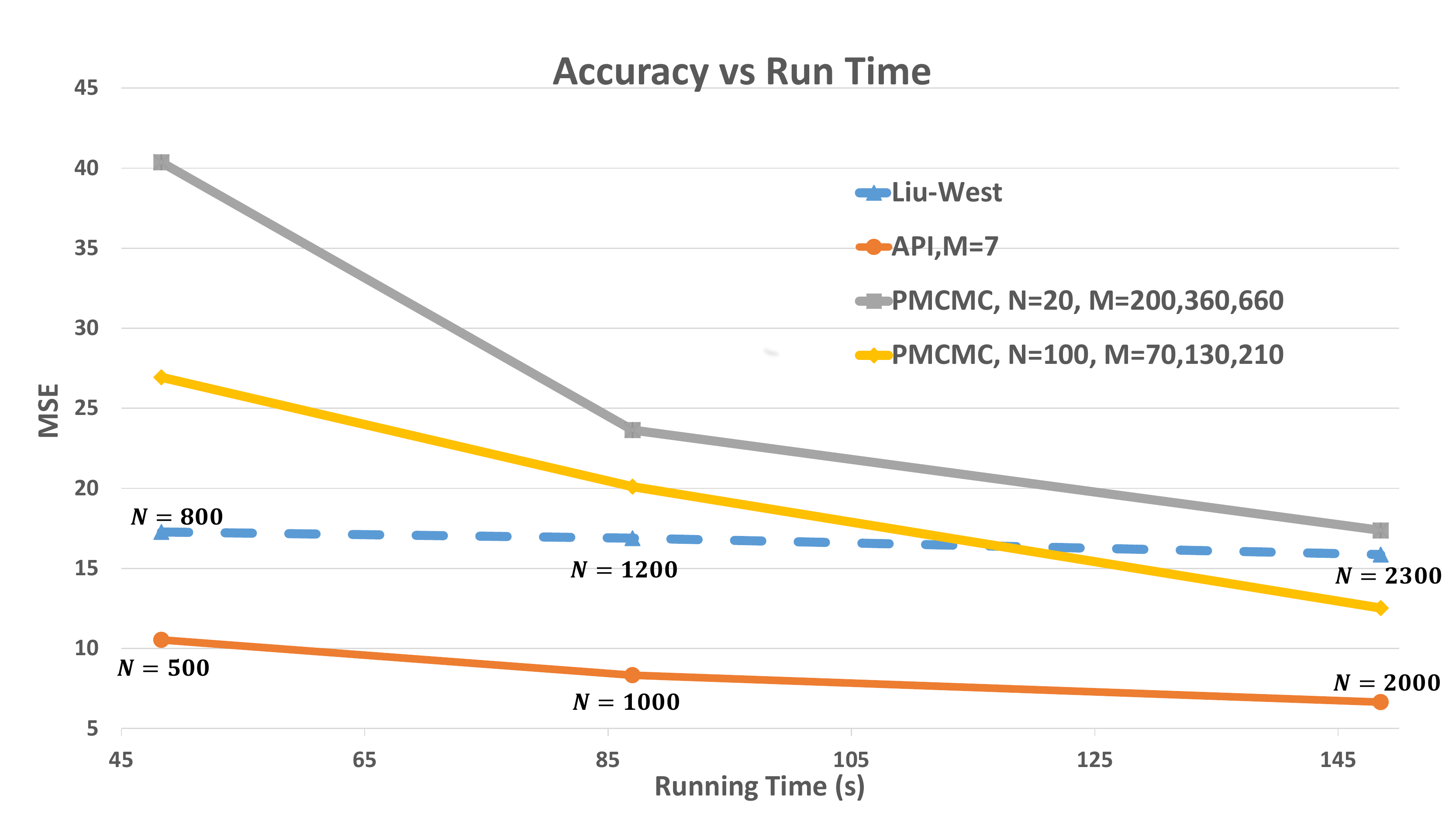}
\vspace{-1.5em}
\caption{Estimation accuracy of \filter against Liu-West filter and PMCMC on the \birdmodel model. \filter produces significantly better estimation results with smaller amount of computation time.}\label{fig:bird_err}
\end{figure}

We again measure the running time of \filter against the plain particle filter on the \birdmodel model. PF uses 104.136s while \filter uses 133.247s for 1000 particles. Note that in this real application,  \filter with $M=7$ only results in 20\% overhead compared with PF, although theoretically \filter should be 28x slower.

\section{Discussion}\label{sec:related}
Similar to \cite{storvik2002particle,lopes2010particle}, we are sampling from $p(\theta \mid x_{0:t}^i,y_{0:t})$ at each time step to fight against sample impoverishment. It has been discussed before that these methods suffer from \emph{ancestral path degeneracy} \cite{chopin2010particle, lopes2010particle, poyiadjis2011particle}. For any number of particles and for a large enough $n$, there exists some $m<n$ such that $p(x_{0:m}\mid y_{0:n})$ is represented by a single unique particle. For dynamic models with long memory, this will lead to a poor approximation of sufficient statistics, which in turn will affect the posterior of the parameters. \citet{poyiadjis2011particle} showed that even under favorable mixing assumptions, the variance of an additive path functional computed via a particle approximation grows quadratically with time. To fight against path degeneracy, one may have to resort to \emph{fixed-lag smoothing} or \emph{smoothing}. \citet{olsson2008sequential} used fixed-lag smoothing to control the variance of the estimates. \citet{del2010forward} proposed an $O(N^2)$ per time step forward smoothing algorithm which leads to variances growing linearly with $t$ instead of quadratically. \citet{poyiadjis2011particle} similarly proposed an $O(N^2)$ algorithm that leads to linearly growing variances. Similarly, doing a full kernel move at a rate of $1/t$ on $\left\{x_{0:t},\theta \right\}$ as in \cite{gilks2001following} will also be beneficial.

Another important matter to consider is the convergence of the assumed density filtering posterior to the true posterior $p(\theta \mid x_{0:t},y_{0:t})$. \citet{opper1998bayesian} analyzed the convergence behavior for the Gaussian projection case. There is no analysis of convergence when the moment matching integrals are computed approximately via Monte Carlo sums. However, our experiments indicate that for approximations with sufficiently many Monte Carlo samples, similar convergence behavior as in \cite{opper1998bayesian} is attained. \citet{heess2013learning} investigated approximating the moment matching integrals robustly and efficiently in the context of expectation-propagation. They train discriminative models that learn to map EP message inputs to outputs. The idea seems promising and can be applied in our setting as well.

% In the context of EP, \citet{minka2005divergence} discussed the effect of minimizing $\alpha$-divergences instead of KL-divergences. Minimizing the KL-divergence leads to an averaging effect whereas for different $\alpha$-divergences differing behavior can be attained. Using different divergence measures can be beneficial especially in the case of multimodal posteriors. 

% Finally, we would like to point to the similarity between our algorithm and another $O(NMT)$ algorithm. For each static parameter particle, we utilize $M$ samples to approximately compute the moment matching integrals. \citet{johansen2012exact} approximated the Rao-Blackwellised particle filters by introducing a top-layer particle filter and a local-level one. The local level particle filter approximates the posterior of the variables to be marginalized by $M$ particles leading to $O(NM)$ computation per time step. 
% Similarly, \citet{chopin2013smc2} combined iterated batch importance sampling of \citet{chopin2002sequential} with particle filtering. For each static parameter particle there is an associated population of state particles leading, again, to $O(NM)$ per time step. A combination of ideas from the aforementioned works and our proposed approach can lead to a non-parametric method that exploits the structure of the dynamic model.

\section{Conclusion}\label{sec:conclude}
In this paper, we present a new inference algorithm, \filter, for both state and parameter estimation in dynamic probabilistic models. We also developed a black-box inference engine, \name, performing \filter over arbitrary models described in the high-level modeling language of \name. \name  leverages multiple compiler level optimizations for efficient computation and achieves 3x to 6x speedup against existing toolboxes. In our experiment, \filter produces order-of-magnitude more accurate estimation result compared to PMCMC within a fixed amount of computation time and is able to handle real-world applications efficiently and accurately.

%One interesting future direction is parallelization since \filter is perfectly parallel. Our tentative results on GPU indicate 10x speedup on  \sinmodel model. Note that memory management becomes more crucial on GPU. How to design an effective parallel back-end remains a challenging problem.

%\newpage
%\input{list_to_be_deleted.tex}
%\newpage
\bibliography{ref}

\begin{thebibliography}{31}
\providecommand{\natexlab}[1]{#1}
\providecommand{\url}[1]{\texttt{#1}}
\expandafter\ifx\csname urlstyle\endcsname\relax
  \providecommand{\doi}[1]{doi: #1}\else
  \providecommand{\doi}{doi: \begingroup \urlstyle{rm}\Url}\fi

\bibitem[Alspach \& Sorenson(1972)Alspach and Sorenson]{alspach1972nonlinear}
Alspach, Daniel~L and Sorenson, Harold~W.
\newblock Nonlinear {B}ayesian estimation using {G}aussian sum approximations.
\newblock \emph{Automatic Control, IEEE Transactions on}, 17\penalty0
  (4):\penalty0 439--448, 1972.

\bibitem[Andrieu et~al.(2010)Andrieu, Doucet, and
  Holenstein]{andrieu2010particle}
Andrieu, Christophe, Doucet, Arnaud, and Holenstein, Roman.
\newblock Particle {M}arkov chain {M}onte {C}arlo methods.
\newblock \emph{Journal of the Royal Statistical Society: Series B (Statistical
  Methodology)}, 72\penalty0 (3):\penalty0 269--342, 2010.

\bibitem[Arora et~al.(2010)Arora, Russell, Kidwell, and
  Sudderth]{arora2010global}
Arora, Nimar~S., Russell, Stuart~J., Kidwell, Paul, and Sudderth, Erik~B.
\newblock Global seismic monitoring as probabilistic inference.
\newblock In \emph{NIPS}, pp.\  73--81, 2010.

\bibitem[Arulampalam et~al.(2002)Arulampalam, Maskell, Gordon, and
  Clapp]{arulampalam2002tutorial}
Arulampalam, Sanjeev, Maskell, Simon, Gordon, Neil, and Clapp, Tim.
\newblock A tutorial on particle filters for on-line non-linear/non-{G}aussian
  {B}ayesian tracking.
\newblock \emph{IEEE Transactions on Signal Processing}, 50\penalty0
  (2):\penalty0 174--188, 2002.

\bibitem[Boyen \& Koller(1998)Boyen and Koller]{boyen1998tractable}
Boyen, Xavier and Koller, Daphne.
\newblock Tractable inference for complex stochastic processes.
\newblock In \emph{Proceedings of the Fourteenth conference on Uncertainty in
  artificial intelligence}, pp.\  33--42. Morgan Kaufmann Publishers Inc.,
  1998.

\bibitem[Capp{\'e} et~al.(2007)Capp{\'e}, Godsill, and
  Moulines]{cappe2007overview}
Capp{\'e}, Olivier, Godsill, Simon~J, and Moulines, Eric.
\newblock An overview of existing methods and recent advances in sequential
  {M}onte {C}arlo.
\newblock \emph{Proceedings of the IEEE}, 95\penalty0 (5):\penalty0 899--924,
  2007.

\bibitem[Chopin et~al.(2010)Chopin, Iacobucci, Marin, Mengersen, Robert, Ryder,
  and Sch{\"a}fer]{chopin2010particle}
Chopin, Nicolas, Iacobucci, Alessandra, Marin, Jean-Michel, Mengersen, Kerrie,
  Robert, Christian~P, Ryder, Robin, and Sch{\"a}fer, Christian.
\newblock On particle learning.
\newblock \emph{arXiv preprint arXiv:1006.0554}, 2010.

\bibitem[Del~Moral et~al.(2010)Del~Moral, Doucet, and Singh]{del2010forward}
Del~Moral, Pierre, Doucet, Arnaud, and Singh, Sumeetpal.
\newblock Forward smoothing using sequential {M}onte {C}arlo.
\newblock \emph{arXiv preprint arXiv:1012.5390}, 2010.

\bibitem[Elmohamed et~al.(2007)Elmohamed, Kozen, and
  Sheldon]{elmohamed2007collective}
Elmohamed, MaS, Kozen, Dexter, and Sheldon, Daniel~R.
\newblock Collective inference on {M}arkov models for modeling bird migration.
\newblock In \emph{Advances in Neural Information Processing Systems}, pp.\
  1321--1328, 2007.

\bibitem[Erol et~al.(2013)Erol, Li, Ramsundar, and Stuart]{erol2013extended}
Erol, Yusuf~B, Li, Lei, Ramsundar, Bharath, and Stuart, Russell.
\newblock The extended parameter filter.
\newblock In \emph{Proceedings of the 30th International Conference on Machine
  Learning (ICML-13)}, pp.\  1103--1111, 2013.

\bibitem[Fearnhead(2002)]{fearnhead2002markov}
Fearnhead, Paul.
\newblock Markov chain {M}onte {C}arlo, sufficient statistics, and particle
  filters.
\newblock \emph{Journal of Computational and Graphical Statistics}, 11\penalty0
  (4):\penalty0 848--862, 2002.

\bibitem[Gilks \& Berzuini(2001)Gilks and Berzuini]{gilks2001following}
Gilks, Walter~R. and Berzuini, Carlo.
\newblock Following a moving target -- {M}onte {C}arlo inference for dynamic
  bayesian models.
\newblock \emph{Journal of the Royal Statistical Society. Series B (Statistical
  Methodology)}, 63\penalty0 (1):\penalty0 127--146, 2001.

\bibitem[Gordon et~al.(1993)Gordon, Salmond, and Smith]{gordon1993novel}
Gordon, N.~J., Salmond, D.~J., and Smith, A. F.~M.
\newblock Novel approach to nonlinear/non-{Gauss}ian {Bayes}ian state
  estimation.
\newblock \emph{IEE Proceedings F Radar and Signal Processing}, 140\penalty0
  (2):\penalty0 107--113, 1993.
\newblock URL \url{http://ieeexplore.ieee.org/stamp/stamp.jsp?arnumber=210672}.

\bibitem[Heess et~al.(2013)Heess, Tarlow, and Winn]{heess2013learning}
Heess, Nicolas, Tarlow, Daniel, and Winn, John.
\newblock Learning to pass expectation propagation messages.
\newblock In \emph{Advances in Neural Information Processing Systems}, pp.\
  3219--3227, 2013.

\bibitem[Huber \& Hanebeck(2008)Huber and Hanebeck]{huber2008gaussian}
Huber, Marco~F and Hanebeck, Uwe~D.
\newblock Gaussian filter based on deterministic sampling for high quality
  nonlinear estimation.
\newblock In \emph{Proceedings of the 17th IFAC World Congress (IFAC 2008)},
  volume~17, 2008.

\bibitem[Julier \& Uhlmann(2004)Julier and Uhlmann]{julier2004unscented}
Julier, Simon~J. and Uhlmann, Jeffrey~K.
\newblock Unscented filtering and nonlinear estimation.
\newblock \emph{Proceedings of the IEEE}, 92\penalty0 (3):\penalty0 401--422,
  2004.

\bibitem[Kantas et~al.(2015)Kantas, Doucet, Singh, Maciejowski, Chopin,
  et~al.]{kantas2015particle}
Kantas, Nikolas, Doucet, Arnaud, Singh, Sumeetpal~S, Maciejowski, Jan, Chopin,
  Nicolas, et~al.
\newblock On particle methods for parameter estimation in state-space models.
\newblock \emph{Statistical science}, 30\penalty0 (3):\penalty0 328--351, 2015.

\bibitem[Lauritzen(1992)]{lauritzen1992propagation}
Lauritzen, Steffen~L.
\newblock Propagation of probabilities, means, and variances in mixed graphical
  association models.
\newblock \emph{Journal of the American Statistical Association}, 87\penalty0
  (420):\penalty0 1098--1108, 1992.

\bibitem[Liu \& West(2001)Liu and West]{liu2001combined}
Liu, Jane and West, Mike.
\newblock Combined parameter and state estimation in simulation-based
  filtering.
\newblock In \emph{Sequential {M}onte {C}arlo Methods in Practice}. 2001.

\bibitem[Lopes \& Tsay(2011)Lopes and Tsay]{lopes2011particle}
Lopes, Hedibert~F and Tsay, Ruey~S.
\newblock Particle filters and bayesian inference in financial econometrics.
\newblock \emph{Journal of Forecasting}, 30\penalty0 (1):\penalty0 168--209,
  2011.

\bibitem[Lopes et~al.(2010)Lopes, Carvalho, Johannes, and
  Polson]{lopes2010particle}
Lopes, Hedibert~F, Carvalho, Carlos~M, Johannes, Michael, and Polson,
  Nicholas~G.
\newblock Particle learning for sequential {B}ayesian computation.
\newblock \emph{Bayesian Statistics}, 9:\penalty0 175--96, 2010.

\bibitem[Milch et~al.(2005)Milch, Marthi, Russell, Sontag, Ong, and
  Kolobov]{milch2005blog}
Milch, Brian, Marthi, Bhaskara, Russell, Stuart, Sontag, David, Ong, Daniel~L.,
  and Kolobov, Andrey.
\newblock Blog: Probabilistic models with unknown objects.
\newblock In \emph{In IJCAI}, pp.\  1352--1359, 2005.

\bibitem[Murphy(2000)]{bayesian2000murphy}
Murphy, Kevin.
\newblock Bayesian map learning in dynamic environments.
\newblock In \emph{In Neural Info. Proc. Systems (NIPS}, pp.\  1015--1021. MIT
  Press, 2000.

\bibitem[Murray(2013)]{murray2013bayesian}
Murray, Lawrence~M.
\newblock Bayesian state-space modelling on high-performance hardware using
  {L}ib{B}i.
\newblock \emph{arXiv preprint arXiv:1306.3277}, 2013.

\bibitem[Olsson et~al.(2008)Olsson, Capp{\'e}, Douc, Moulines,
  et~al.]{olsson2008sequential}
Olsson, Jimmy, Capp{\'e}, Olivier, Douc, Randal, Moulines, Eric, et~al.
\newblock Sequential {M}onte {C}arlo smoothing with application to parameter
  estimation in nonlinear state space models.
\newblock \emph{Bernoulli}, 14\penalty0 (1):\penalty0 155--179, 2008.

\bibitem[Opper \& Winther(1998)Opper and Winther]{opper1998bayesian}
Opper, Manfred and Winther, Ole.
\newblock A {B}ayesian approach to on-line learning.
\newblock \emph{On-line Learning in Neural Networks, ed. D. Saad}, pp.\
  363--378, 1998.

\bibitem[Poyiadjis et~al.(2011)Poyiadjis, Doucet, and
  Singh]{poyiadjis2011particle}
Poyiadjis, George, Doucet, Arnaud, and Singh, Sumeetpal~Sindhu.
\newblock Particle approximations of the score and observed information matrix
  in state space models with application to parameter estimation.
\newblock \emph{Biometrika}, 98\penalty0 (1):\penalty0 pp. 65--80, 2011.
\newblock ISSN 00063444.
\newblock URL \url{http://www.jstor.org/stable/29777165}.

\bibitem[Seeger(2005)]{expectation05seeger}
Seeger, Matthias.
\newblock Expectation propagation for exponential families.
\newblock Technical report, 2005.

\bibitem[Storvik(2002)]{storvik2002particle}
Storvik, Geir.
\newblock Particle filters for state-space models with the presence of unknown
  static parameters.
\newblock \emph{IEEE Transactions on Signal Processing}, 50\penalty0
  (2):\penalty0 281--289, 2002.

\bibitem[Todeschini et~al.(2014)Todeschini, Caron, Fuentes, Legrand, and
  Del~Moral]{todeschini2014biips}
Todeschini, Adrien, Caron, Fran{\c{c}}ois, Fuentes, Marc, Legrand, Pierrick,
  and Del~Moral, Pierre.
\newblock {B}iips: Software for {B}ayesian inference with interacting particle
  systems.
\newblock \emph{arXiv preprint arXiv:1412.3779}, 2014.

\bibitem[Zoeter \& Heskes(2005)Zoeter and Heskes]{zoeter04a}
Zoeter, Onno and Heskes, Tom.
\newblock Gaussian quadrature based expectation propagation.
\newblock In Ghahramani, Z. and Cowell, R. (eds.), \emph{Proceedings of the
  Tenth International Workshop on Artificial Intelligence and Statistics}, pp.\
   445--452. Society for Artificial Intelligence and Statistics, 2005.

\end{thebibliography}
\bibliographystyle{icml2016}

\newpage

%\clearpage
%\appendix
%\appendixpage

%\twocolumn[
%\maketitle
%\begin{center}
%\LARGE{Supplemental Material for\\ Towards Practical  Parameter and State Estimation}
%\vspace{3em}
%\end{center}
%]

%\input{appendix.tex}

\end{document}